\documentclass{article} 
\usepackage{iclr2026_conference,times}

\usepackage{amsmath,amsfonts,bm}

\def\eqref#1{equation~\ref{#1}}

\def\1{\bm{1}}

\DeclareMathAlphabet{\mathsfit}{\encodingdefault}{\sfdefault}{m}{sl}
\SetMathAlphabet{\mathsfit}{bold}{\encodingdefault}{\sfdefault}{bx}{n}

\usepackage{hyperref}       
\usepackage{url}            
\usepackage{booktabs}       
\usepackage{amsfonts}       
\usepackage{nicefrac}       
\usepackage{microtype}      
\usepackage{amsmath, amssymb, amsthm}
\usepackage{colortbl}
\usepackage{pifont} 
\usepackage{multirow}
\usepackage{multicol}
\usepackage{makecell}
\usepackage{graphicx}
\usepackage{diagbox}
\usepackage{subcaption}
\usepackage[utf8]{inputenc}
\usepackage{svg}
\usepackage{comment} 

\usepackage{float}
\usepackage{enumitem}
\usepackage{caption}
\usepackage{graphicx}
\usepackage{caption}
\usepackage{booktabs}
\usepackage{graphicx}    
\usepackage{caption}     
\usepackage{subcaption}  
\usepackage{newunicodechar}
\newunicodechar{，}{,}
\title{FreqCa: Accelerating Diffusion Models via Frequency-Aware Caching}

\vspace{-10mm}
\author{Jiacheng Liu$^{1,2,3}$\thanks{Equal contribution.} \,
  Peiliang Cai$^{1\,*}$ \,
  Qinming Zhou$^{1,4}$ \,
  Yuqi Lin$^{1,5}$ \,
  Deyang Kong$^{1，7}$ \\
  \textbf{Benhao Huang}$^{1,6}$ \,
  \textbf{Yupei Pan}$^{1,7}$ \,
  \textbf{Haowen Xu}$^{1}$ \,
  \textbf{Chang Zou}$^{1,2,7}$ \,
  \textbf{Junshu Tang}$^{2}$ \\
  \textbf{Shikang Zheng}$^{1,8}$ \,
  \textbf{Linfeng Zhang}$^{1}$\thanks{Corresponding author. Email: \texttt{zhanglinfeng@sjtu.edu.cn}} \\
  \vspace{4mm}
  $^{1}$EPIC Lab,STJU \quad
  $^{2}$Tencent Hunyuan\quad
  $^{3}$SDU\quad
  $^{4}$THU\quad
  $^{5}$JLU\quad
  $^{6}$CMU\quad
  $^{7}$UESTC\quad
  $^{8}$SCUT\quad
}

\iclrfinalcopy 
\begin{document}

\maketitle
\vspace{-6mm}
\begin{abstract}
\vspace{-4mm}
The application of diffusion transformers is suffering from their significant inference costs. Recently, feature caching has been proposed to solve this problem by reusing features from previous timesteps, thereby skipping computation in future timesteps. 
However, previous feature caching assumes that features in adjacent timesteps are similar or continuous, which does not always hold in all settings.
To investigate this, this paper begins with an analysis from the frequency domain, which reveal that \emph{different frequency bands in the features of diffusion models exhibit different dynamics across timesteps.} Concretely, low-frequency components, which decide the structure of images, exhibit higher \emph{similarity} but poor continuity. In contrast, the high-frequency bands, which decode the details of images, show significant continuity but poor similarity. These interesting observations motivate us to propose  \textbf{Freq}uency-aware \textbf{Ca}ching (\textbf{FreqCa})
 which directly reuses features of low-frequency components based on their similarity, while using a second-order Hermite interpolator to predict the volatile high-frequency ones based on its continuity.
 Besides, we further propose to cache Cumulative Residual Feature (CRF) instead of the features in all the layers, which reduces the memory footprint of feature caching by \textbf{99\%}. 
 Extensive experiments on FLUX.1-dev, FLUX.1-Kontext-dev, Qwen-Image, and Qwen-Image-Edit demonstrate its effectiveness in both generation and editing. \emph{Codes are available in the supplementary materials and will be released on GitHub.}

\end{abstract}
\vspace{-4mm}

\section {Introduction}

\begin{figure}[htp]
    \centering
    \includegraphics[width=0.9\linewidth]{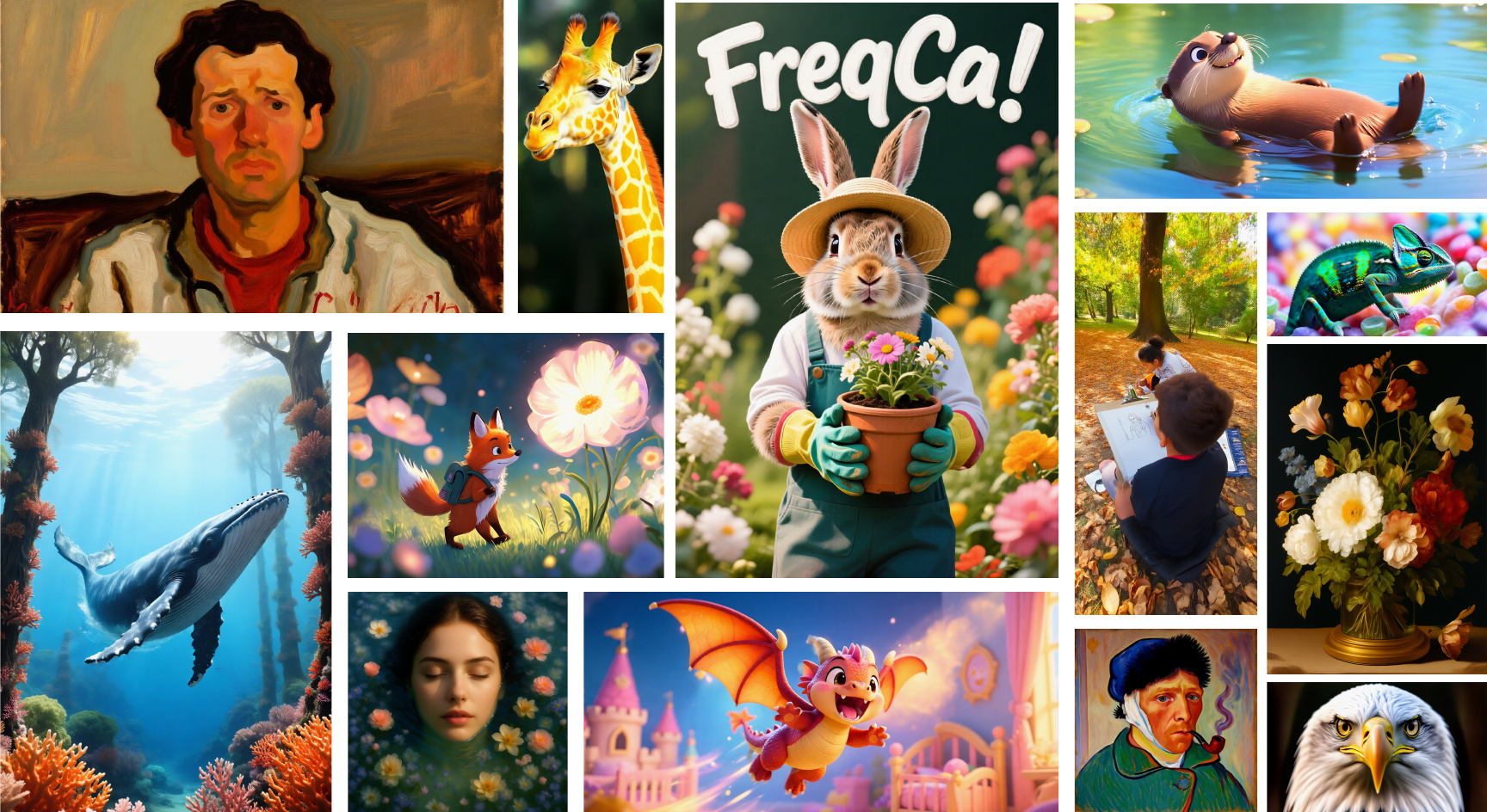}
    \caption{Images sampled by Qwen-image with \textit{FreqCa} with 7.14$\times$ acceleration. }
    \label{fig:showcase}
\end{figure}
\vspace{-3mm}   

\begin{figure}[htbp]
    \centering
    \includegraphics[width=\linewidth]
    {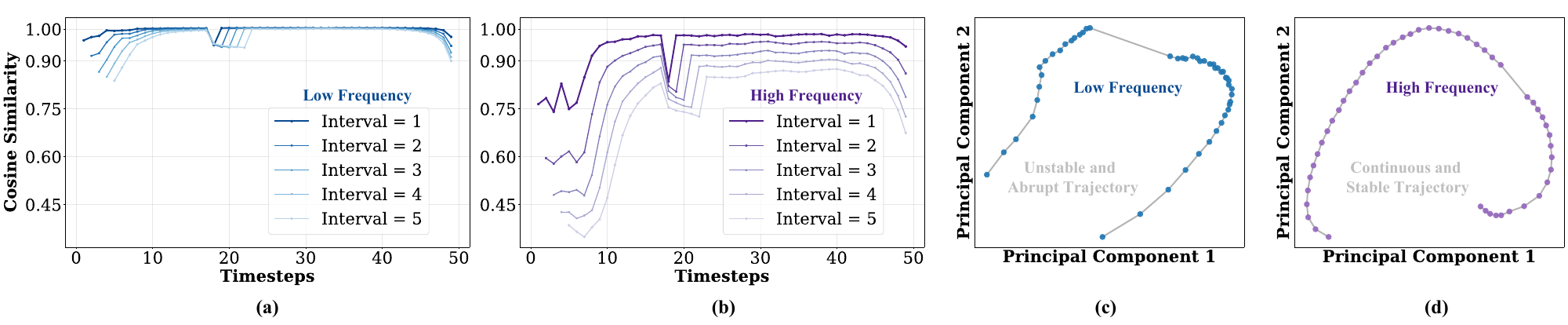}
        \vspace{-6mm}
    \captionof{figure}{\textbf{Analysis from the frequency perspective.} \textbf{(a)-(b)}:Temporal similarity analysis using cosine similarity for low-frequency and high-frequency components across different step intervals. \textbf{(c)-(d)}: Feature trajectory visualized via Principal Component Analysis (PCA). }
    \label{fig:analysis}
    \vspace{-7mm}
\end{figure}

Diffusion Models (DMs) have achieved remarkable success in generative tasks like image synthesis and video generation \citep{DM,StableDiffusion,blattmann2023SVD}. The recent introduction of Diffusion Transformers \citep{DiT} has further advanced generation quality and diversity, establishing them as the predominant architecture for large-scale visual content creation. However, diffusion transformers typically rely on a stack of heavy transformer blocks and multi-step sampling, making computational efficiency a critical bottleneck for their practical deployment. 
To address this, the paradigm of feature caching has emerged, which exploits the high temporal redundancy between adjacent timesteps for acceleration~\citep{ma2024deepcache,li2023FasterDiffusion,selvaraju2024fora,chen2024delta-dit,zou2024accelerating,zou2024DuCa}.

\textbf{The debate of caching paradigms.} Feature caching has gradually emerged into two different paradigms. The paradigm of ``Cache-Then-Reuse'' assumes that the features of DM in adjacent timesteps are highly \textbf{similar} and thus proposes to directly \textbf{reuse} the features of previous timesteps in the future timesteps~\citep{selvaraju2024fora}.  In contrast, the paradigm of ``Cache-Then-Forecast'' assumes that features of DM are ``continuous'' and thus proposes to forecast features in the future timesteps based on features in the previous timesteps with non-parametric predictors such as Taylor expansion. Although the paradigm ``Cache-Then-Forecast''    tends to show better performance in recent works, their assumption of continuity does not always hold perfectly. For instance, Liu~\emph{et al.} demonstrates that features of FLUX are not high-order continuous, making TaylorSeer degenerate into a linear prediction method and thus suffer from quality loss~\citep{liuTaylorSeer2025}. Based on these findings, this paper begins with an in-depth analysis of the temporal dynamics of diffusion models. 

\textbf{Analysis from the frequency perspective.} In classical image processing, the high-frequency and low-frequency components of images are usually considered as carrying different semantic information,  which motivates us to study the dynamics of high-frequency and low-frequency components in the feature of diffusion models separately. As shown in Figure~\ref{fig:analysis}(a)-(b), surprisingly, we find different frequency exhibits significantly different dynamics. Concretely, the similarity of low-frequency is higher than \emph{0.90} at most timesteps, while the high-frequency exhibits clearly low similarity.
On the other hand, as shown in Figure~\ref{fig:analysis}(c)-(d), the feature trajectory of high-frequency shows perfect stability and continuity, while the feature trajectory of low-frequency is unstable and accompanied by sudden mutation, indicating that the high-frequency information can be accurately predicted while the low-frequency information fails.

Based on the above observations, this paper introduces \textbf{Freq}uency-aware Feature \textbf{Ca}ching (\textbf{FreqCa}), which aims to decouple the frequency of the features in diffusion models and treat them in different paradigms. Concretely, \textit{FreqCa} applies any frequency decomposition (\emph{e.g, Fourier Transformation}) to the cached features. For the low-frequency bands, we directly reuse them in the future timesteps because of their \emph{high similarity}. For the high-frequency bands, we predict their values in the future timesteps with any sequential predictor (\emph{Hermite polynomial predictor}) for their \emph{good continuity}. Then, in the future timesteps, \textit{FreqCa} reconstructs the features based on the reused low-frequency bands and predicted high-frequency bands,
enabling it to skip the computation over diffusion transformers, achieving the best cooperation between the previous caching paradigms.

\textbf{Memory-Efficient Feature Caching.}
The previous caching methods usually cache all the features from attention and FFN layers, leading to significant memory costs (\emph{e.g,} $\geq$ 10G memory costs on FLUX in ToCa), preventing feature caching methods from real-world applications. As discussed by \cite{veit2016residual}, neural networks with residual connections can be considered as an ensemble of features in all blocks, which motivates us to propose caching the Cumulative Residual Feature (CRF), \emph{i.e.,} the cumulative features of all the residual connections from attention and FFN blocks. This trick helps us reduce the memory footprint from caching $2\times L$ ($L$ indicates layer counts) features into a single feature vector,  slashing cache memory usage by up to 99\%. Besides, it also reduces the number of frequency (reverse) decomposition operations by $2L$ times, making them account for only $\leq0.01\%$ latency costs during the whole diffusion process.

In summary, this contribution of this paper is as follows.



\begin{itemize}[leftmargin=10pt,topsep=-2pt]

    \item \textbf{Frequency-Aware Feature Caching:} Motivated by the difference in similarity and continuity of different frequency bands, we propose \textit{FreqCa}, which applies different feature caching methods to different frequencies, unifying the two previous caching paradigms.
    \item \textbf{Memory-Efficient Feature Caching } By caching only the Cumulative Residual Feature, \textit{FreqCa} achieves \textbf{$\mathcal{O}(1)$} memory complexity, slashing Cache memory usage to a mere \textbf{1\%} of prior approaches without fidelity loss, enabling high-quality acceleration on consumer hardware.
    \item \textbf{State-of-the-art generalization and performance:} Across text-to-image generation and image editing tasks, \textit{FreqCa} consistently delivers 6–7$\times$ acceleration with quality degradation below 2\%, outperforming existing methods and demonstrating strong robustness and practicality.
\end{itemize}

\vspace{-2mm}

\section{Related Works}\label{sec:Related Works}
\vspace{-3mm}

Diffusion models have emerged as a cornerstone of modern generative AI, exhibiting state-of-the-art capabilities in synthesizing visual content~\citep{sohl2015deep,ho2020DDPM}. While early models were predominantly built upon U-Net architectures~\citep{ronneberger2015unet}, their scalability limitations paved the way for the Diffusion Transformer (DiT)~\citep{peebles2023dit}. The DiT architecture has since become foundational, catalyzing a wave of powerful models across diverse domains~\citep{opensora,yang2025cogvideox}. Nevertheless, the iterative nature of the diffusion sampling process imposes a significant computational burden during inference, making acceleration a critical area of research~\citep{ho2020DDPM,peebles2023dit}. Current efforts to enhance efficiency are largely focused on two complementary directions: reducing the number of sampling steps and accelerating the denoising network itself.

\vspace{-3mm}
\subsection{Sampling Timestep Reduction}
\vspace{-1mm}

One primary strategy seeks to minimize the number of required sampling iterations while preserving generation quality. Seminal work like DDIM introduced deterministic sampling to reduce step counts without significant fidelity loss~\citep{songDDIM}. This concept was further refined by the DPM-Solver series, which employed high-order ODE solvers to achieve faster convergence~\citep{lu2022dpm,lu2022dpm++,zheng2023dpmsolvervF}. Other notable approaches include knowledge distillation, which trains a student model to emulate multiple denoising steps of a larger teacher model~\citep{salimans2022progressive,meng2022on}, and Rectified Flow, which learns to straighten the generation path between noise and data distributions~\citep{refitiedflow}. More recently, Consistency Models have enabled high-quality synthesis in a single step by directly mapping noise to clean data, circumventing the need for a sequential path~\citep{song2023consistency}.

\vspace{-3mm}
\subsection{Denoising Network Acceleration}
\vspace{-1mm}

An alternative to reducing timesteps is to decrease the computational cost of each forward pass through the denoising network. This is typically achieved via model compression or feature caching.

\vspace{-2mm}
\paragraph{Model Compression-based Acceleration.} 
\vspace{-1mm}

One avenue involves model compression, which includes techniques such as network pruning~\citep{structural_pruning_diffusion, zhu2024dipgo}, quantization~\citep{10377259, shang2023post, kim2025ditto}, and various forms of token reduction that dynamically shorten the input sequence length~\citep{bolya2023tomesd, kim2024tofu, zhang2024tokenpruningcachingbetter, zhang2025sito}. While effective, these methods often necessitate a fine-tuning or retraining stage to mitigate the potential loss of expressive power inherent in model simplification~\citep{li2024snapfusion,10377259}.

\vspace{-2mm}
\paragraph{Feature Caching-based Acceleration.}
\vspace{-2mm}
A compelling training-free alternative is feature caching, which exploits temporal redundancies in the denoising process. Pioneered in U-Net architectures through FasterDiffusion and DeepCache, this paradigm was subsequently adapted to DiTs. Initial efforts focused on a ``cache then reuse'' strategy, while advanced techniques like FORA and $\Delta$-DiT refined this approach. This concept evolved with more sophisticated mechanisms, including dynamic token-level updates (ToCa), adaptive sampling (RAS~\citep{liu2025regionadaptivesamplingdiffusiontransformers}), and explicit error correction frameworks~\citep{qiu2025acceleratingdiffusiontransformererroroptimized, chenIncrementCalibrated2025, chuOmniCache2025}. A pivotal shift was the ``cache then forecast'' paradigm introduced by TaylorSeeer, which was further advanced by more robust numerical methods in FoCa~\citep{zhengFoCa2025}, HiCache~\citep{fengHiCache2025}, and SpeCa~\citep{Liu2025SpeCa}.

However, a crucial flaw underlies these sophisticated paradigms, as hinted at by preliminary frequency-domain analyses. For instance, PAB~\citep{zhao2024PAB} insightfully associated different attention mechanisms with distinct frequency bands but did not delve into token-level frequency dynamics. Similarly, while FasterCache~\citep{lvFasterCacheTrainingFreeVideo2025} examined the frequency-domain differences within Classifier-Free Guidance , its findings were confined to this specific context, not addressing the more universal dynamics of temporal feature evolution and thus showing limited practical acceleration.

In contrast to prior methods that treat features as a monolithic whole, we propose \textbf{FreqCa}, which resolves quality degradation in caching by decomposing features into their stable low-frequency and volatile high-frequency components for differentiated treatment. As an added benefit,  we introduce the Cumulative Residual Feature, collapsing the memory complexity from $\mathcal{O}(L)$ to $\mathcal{O}(1)$ to solve the resource inefficiency of prior ``layer-wise'' architectures.
\vspace{-4mm}

\section{Method}
\vspace{-1.5mm}
\subsection{Preliminary}

\vspace{-1.5mm}
\subsubsection{Diffusion Transformer Architecture.}  
\vspace{-1.5mm}
The Diffusion Transformer (DiT)~\citep{DiT} employs a hierarchical structure 
$\mathcal{G} = g_1 \circ g_2 \circ \cdots \circ g_L$, 
where each module $g_l = \mathcal{F}_{\text{SA}}^l \circ \mathcal{F}_{\text{CA}}^l \circ \mathcal{F}_{\text{MLP}}^l$ 
is composed of self-attention (SA), cross-attention (CA), and multilayer perceptron (MLP) components. 
In DiT, these components are dynamically adapted over time to handle different noise levels during the image generation process. 
The input $\mathbf{x}_t = \{x_i\}_{i=1}^{H \times W}$ is represented as a sequence of tokens corresponding to image patches. 
Each module integrates information through residual connections of the form 
$\mathcal{F}(\mathbf{x}) = \mathbf{x} + \text{AdaLN} \circ f(\mathbf{x})$, 
where AdaLN denotes adaptive layer normalization, which stabilizes training and improves learning effectiveness.

\vspace{-1.5mm}
\subsubsection{Frequency Decomposition Methods }
\vspace{-1mm}
Frequency decomposition, through methods like the Fast Fourier Transform (FFT) and Discrete Cosine Transform (DCT), is a powerful technique for decoupling signals into distinct components. This process separates a signal into its \textbf{low-frequency} components, which typically represent global structure and smooth layouts, and its \textbf{high-frequency} components, which correspond to fine-grained details and sharp edges. In the context of diffusion models, this decoupling allows us to differentiate between stable, foundational structures and volatile, transient details along the generative trajectory.

\vspace{-2mm}
\subsection{Frequency-Aware Cache Acceleration Framework}
\vspace{-2mm}
In this section, we introduce the \textbf{FreqCa (Frequency-aware Feature Caching)} framework, which is built upon three key components: (i) performing frequency decomposition on the feature to be cached and applying separate strategies for its low- and high-frequency components;  (ii) employing a nonlinear Hermite-polynomial-based predictor for the high-frequency part to improve prediction accuracy; and (iii) identifying the Cumulative Residual Feature (CRF) as a novel, highly efficient single-tensor caching target that encapsulates the entire transformation history of the model.

\begin{figure}[htb]
    \centering
    \includegraphics[width=1\linewidth]{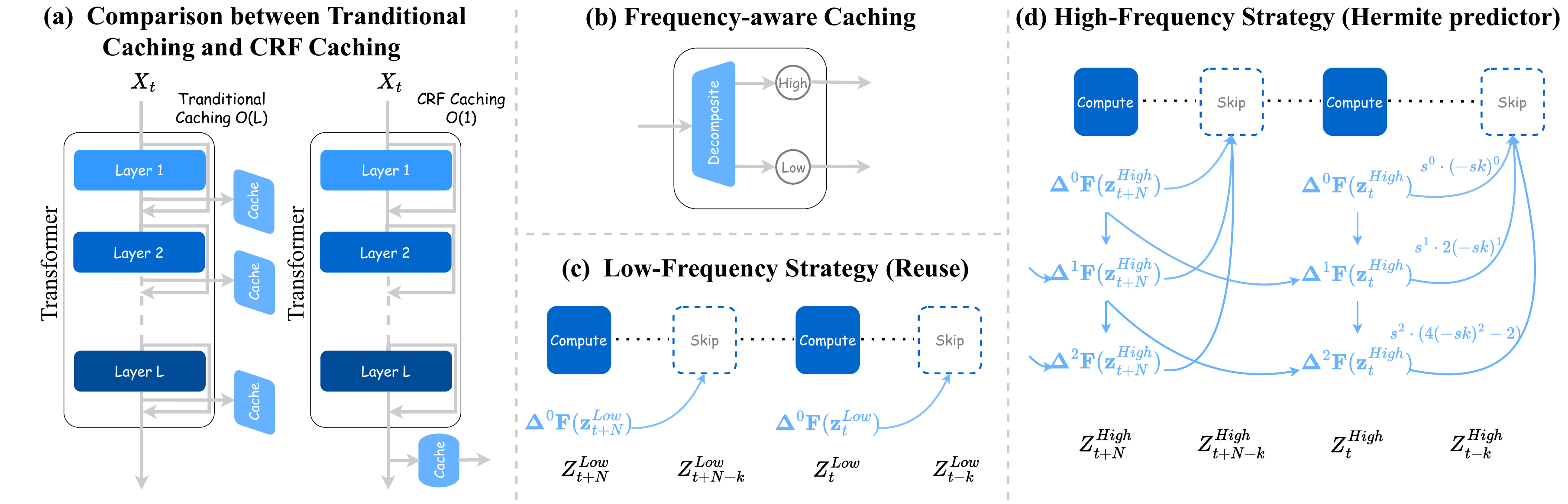}
    \vspace{-7mm}
    \caption{\textbf{Overview of the FreqCa framework.}\textbf{ (a) CRF Caching }: Instead of caching features at every layer, we cache only the single Cumulative Residual Feature (CRF) at the end. \textbf{(b) Frequency-aware Caching}: The cached features are separated into low- and high-frequency bands using frequency decomposition techniques such as FFT or DCT. \textbf{(c) Low-Frequency Strategy}: Low-frequency component is directly reused from the prior step. \textbf{(d) High-Frequency Strategy}:  High-frequency component is forecasted using a Hermite predictor fitted on the last two activated steps.}
    \label{fig:method}
    \vspace{-2mm}
\end{figure}

\vspace{-2.5mm}
\paragraph{1. Frequency-Decomposed Caching and Prediction Strategy.} 

Our differentiated caching strategy is motivated by the distinct temporal dynamics of frequency components. Low-frequency components exhibit high similarity but low continuity, making them stable but difficult to predict. Conversely, high-frequency components are less similar but more continuous, making them volatile yet predictable along a trajectory. This key difference means that a one-size-fits-all approach is not the best and that a differentiated strategy is required.

To implement this, we first decompose the feature $\mathbf{z}_t$ into its constituent parts using a generic frequency transform $\mathcal{D}(\cdot)$:
\[
\mathbf{z}_t = \mathbf{z}_t^{\text{low}} + \mathbf{z}_t^{\text{high}}, \quad \text{where} \quad \mathbf{z}_t^{\text{low/high}} = \mathcal{P}_{\text{low/high}}\!\big(\mathcal{D}(\mathbf{z}_t)\big).
\]
Here, $\mathcal{P}_{\text{low/high}}$ are complementary projection operators. The low-frequency part governs global structure, while the high-frequency part encodes fine details.

Based on their dynamics, we apply tailored strategies. For \textbf{the stable low-frequency component} $\mathbf{z}_t^{\text{low}}$, we apply a direct reuse strategy to maintain global consistency with negligible cost:
$
    \widehat{\mathbf{z}}_t^{\text{low}} = \mathbf{z}_{t-1}^{\text{low}}.
$

For the \textbf{predictable high-frequency component} $\mathbf{z}_t^{\text{high}}$, we employ a nonlinear predictor based on Hermite polynomials to accurately forecast its trajectory. Each high-frequency coefficient $\widehat{h}_i$ at a normalized time $s_t \in [-1,1]$ is modeled as:
$
    \widehat{h}_i(s_t) \;=\; \sum_{k=0}^{m} c_{i,k}\,\mathrm{He}_k(s_t),
$
where the coefficients $c_{i,k}$ are estimated via least-squares regression from the $K$ most recent cached steps. This yields the precisely reconstructed high-frequency component $\widehat{\mathbf{z}}_t^{\text{high}}$.

Finally, the two components are recombined  to yield the final predicted feature, $\widehat{\mathbf{z}}t = \widehat{\mathbf{z}}_t^{\text{low}} + \widehat{\mathbf{z}}_t^{\text{high}}$.

\vspace{-2mm}
\paragraph{2.Cumulative Residual Feature (CRF)} 
At its core, a Diffusion Transformer (DiT) is a deep stack of $L$ residual blocks. The transformation at each block $l$ is not a replacement but an incremental update, as described by the standard residual connection:
$
    \mathbf{h}^{(l+1)} \;=\; \mathbf{h}^{(l)} \;+\; \mathcal{F}^{(l)}\!\big(\mathbf{h}^{(l)},\, t\big),
$
where $\mathcal{F}^{(l)}(\cdot,t)$ denotes the transformation module at layer $l$ (including Attention, and MLP), which is dynamically modulated by the diffusion timestep $t$ (e.g., through AdaLN).

The structure of the DiT's final output is thus revealed: $\phi_L(\mathbf{x}_t) \;=\; \mathbf{h}^{(0)} \;+\; \sum_{l=0}^{L-1} \mathcal{F}^{(l)}\!\big(\mathbf{h}^{(l)},\, t\big)$.
This formulation shows that the final output is not just another intermediate feature, but the \textbf{accumulation of the initial input and all subsequent residual updates}. We define this special output $\mathbf{z}_t \;\triangleq\; \phi_L(\mathbf{x}_t)$ and name it the \textbf{Cumulative Residual Feature (CRF)}, reflecting its composite nature.

\begin{figure}[htbp]
  \centering
  \begin{minipage}[t]{0.65\textwidth}
    \centering
       \raisebox{5mm}{\includegraphics[width=1.09\linewidth]{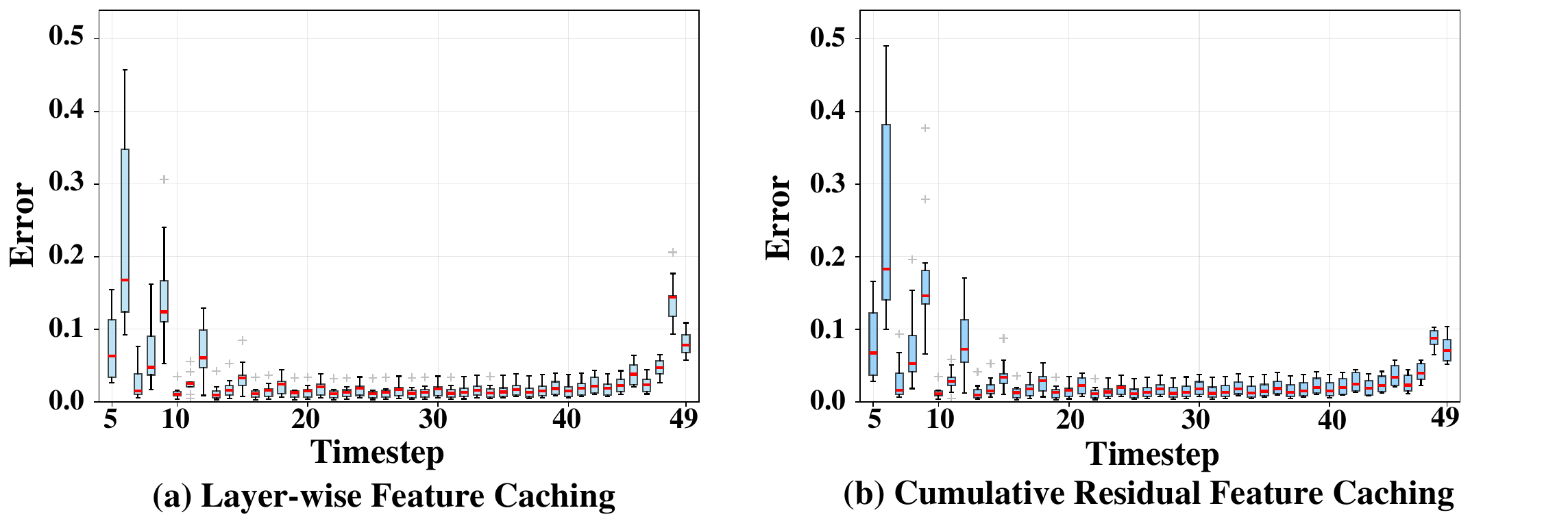}}
       \vspace{-9mm}
    \caption{Box plots of Mean Squared Error (MSE) between ground-truth and predicted features per timestep. (a) layer-wise feature caching and (b) cumulative residual feature (CRF) caching.}
    \label{fig:lite_error_analysis}
    \vspace{-8mm}
  \end{minipage}\hfill%
  \begin{minipage}[t]{0.32\textwidth}
    \centering
    \includegraphics[width=1.095\linewidth]{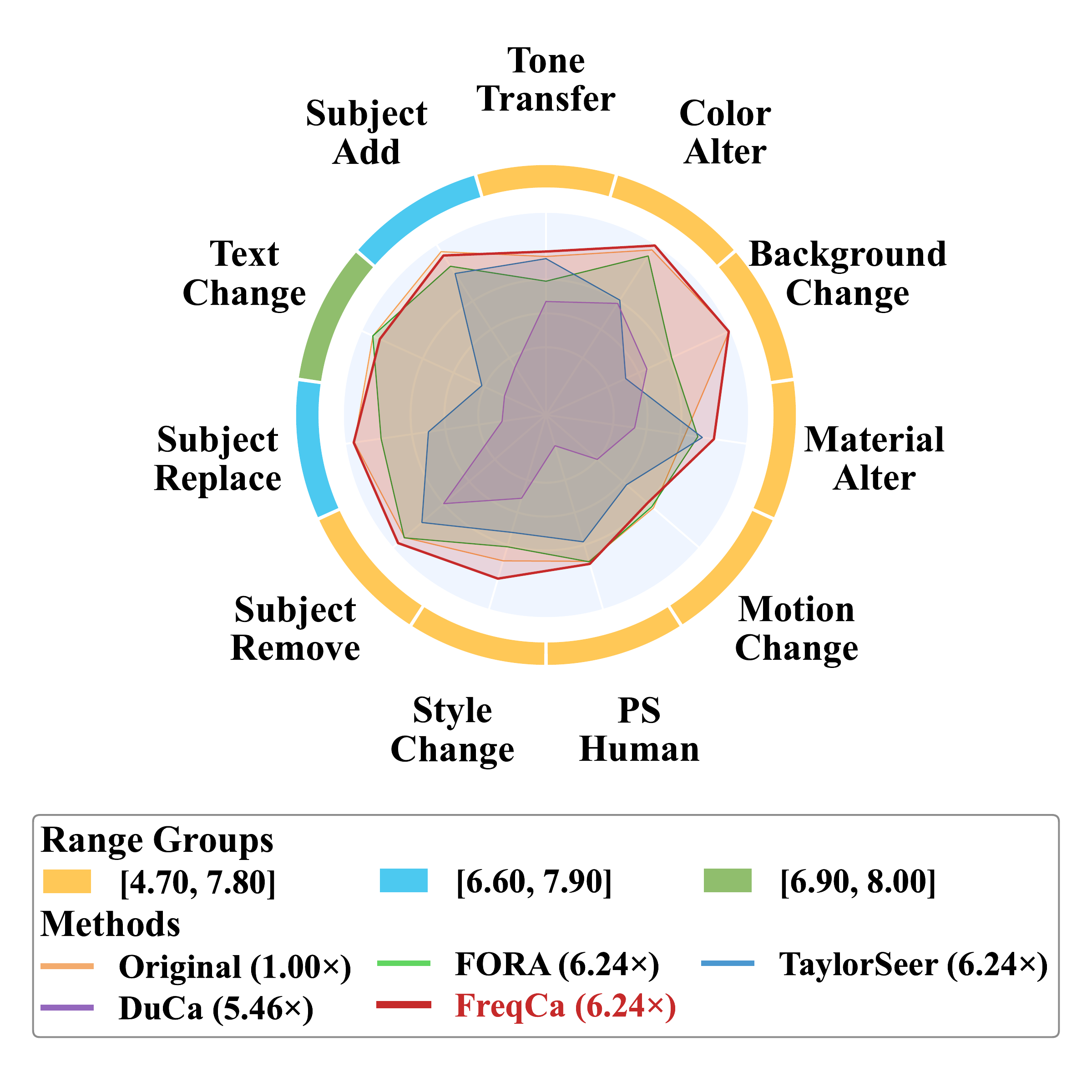}
    \vspace{-9mm}
    \captionof{figure}{Gedit Benchmark on Qwen-Image-Edit, {\textit{FreqCa}} outperforms most baselines.}
    \label{fig:quality}
    \vspace{-8mm}
  \end{minipage}
\end{figure}

This insight leads to a more memory-efficient strategy. While conventional layer-wise caching must store all intermediate features $\{\mathbf{h}^{(l)}\}_{l=0}^{L-1}$, our approach uses the fact that the CRF already contains the entire transformation history. We use this single, globally fused tensor as a highly efficient replacement for the full feature set. As shown in Figure~\ref{fig:lite_error_analysis}, caching only the CRF achieves nearly identical reconstruction fidelity to full layer-wise caching, incurring only a 4\% higher MSE on average, which confirms that the CRF acts as a near-lossless compression of the entire computational path. This makes it an ideal lightweight caching target, enabling a revolutionary reduction in memory complexity from $\mathcal{O}(L)$ to $\mathcal{O}(1)$ without a meaningful sacrifice in quality.

\vspace{-8mm}

\vspace{-1.5mm}

\section {Experiment}
\vspace{-1.5mm}

\subsection{Experiment Settings}
\vspace{-1.5mm}

\paragraph{Model Configurations.}
The experiments are conducted on four state-of-the-art visual generative models—\textbf{FLUX.1-dev}~\citep{flux2024}, \textbf{Qwen-Image}~\citep{liu2023rectified}, \textbf{
FLUX.1-Kontext-dev}~\citep{kontext2025}, and \textbf{Qwen-Image-Edit}~\citep{salimans2022progressive}.

\vspace{-1.5mm}
\paragraph{Evaluation and Metrics.} 
For the text-to-image generation evaluation, we adopt the DrawBench~\citep{saharia2022drawbench} benchmark. The generated samples are systematically evaluated using ImageReward~\citep{xu2023imagereward} and CLIP Score~\citep{hessel2021clipscore}, which jointly measure image quality and text–image semantic alignment. To assess visual fidelity, we further employ PSNR, SSIM~\citep{wang2004imagequality} and LPIPS~\citep{zhangUnreasonableEffectivenessDeep2018}, thereby capturing both pixel-level similarity and perceptual consistency. Additionally, we evaluate general-purpose image editing using the GEdit benchmark~\citep{gedit2024}, which systematically assesses instruction-driven editing fidelity and alignment to target modifications under textual and visual guidance.

\vspace{-1.5mm}
\subsection{Text-to-Image Generation}
\vspace{-1.5mm}

\subsubsection{FLUX.1-dev}
\vspace{-1.5mm}
\begin{table*}[hbt]
\centering
\caption{\textbf{Quantitative comparison in text-to-image generation} for FLUX.1-dev and FLUX.1-schnell(a distilled version). Best results are highlighted in \textbf{bold}, and second-best are \underline{underlined}.}
\vspace{-3mm}
  \resizebox{\textwidth}{!}{ 
    \begin{tabular}{l | c  c | c  c | c  c | c c c}
        \toprule
        \multirow{2}{*}{\textbf{Method}}
        & \multicolumn{4}{c|}{\textbf{Acceleration}} 
        & \multicolumn{2}{c|}{\textbf{Quality Metrics}} 
        & \multicolumn{3}{c}{\textbf{Perceptual Metrics}}\rule{0pt}{2ex}\\
        \cline{2-10}
        & \textbf{Latency(s) \(\downarrow\)} 
        & \textbf{Speed \(\uparrow\)} 
        & \textbf{FLOPs(T) \(\downarrow\)}  
        & \textbf{Speed \(\uparrow\)} 
        & \textbf{ImageReward\(\uparrow\)} 
        & \textbf{CLIP\(\uparrow\)} 
        & \textbf{PSNR\(\uparrow\)} 
        & \textbf{SSIM\(\uparrow\)}
        & \textbf{LPIPS\(\downarrow\)}\rule{0pt}{2ex}\\
        \midrule

\textbf{[dev]: 50 steps}
& 23.24 \textcolor{gray!70}{\scriptsize (+0.0\%)} & 1.00$\times$ & 3726.87 & 1.00$\times$ & 0.99 \textcolor{gray!70}{\scriptsize (+0.0\%)} & 32.64 \textcolor{gray!70}{\scriptsize (+0.0\%)} & $\infty$ & 1.00 & 0.00  \\

\textbf{$60\%$ steps}
& 14.12 \textcolor{gray!70}{\scriptsize (-39.2\%)} & 1.65$\times$ & 2236.12 & 1.67$\times$ & 0.97 \textcolor{gray!70}{\scriptsize (-2.0\%)} & 32.66 \textcolor{gray!70}{\scriptsize (+0.1\%)} & 30.31 & 0.78 & 0.25  \\

\textbf{$50\%$ steps}
& 11.82 \textcolor{gray!70}{\scriptsize (-49.1\%)} & 1.97$\times$ & 1863.44 & 2.00$\times$ & 0.97 \textcolor{gray!70}{\scriptsize (-2.0\%)} & 32.57 \textcolor{gray!70}{\scriptsize (-0.2\%)} & 29.56 & 0.73 & 0.31  \\

\textbf{PAB}
& 17.84 \textcolor{gray!70}{\scriptsize (-23.2\%)} & 1.30$\times$ & 3013.13 & 1.24$\times$ &0.95 \textcolor{gray!70}{\scriptsize (-4.0\%)} & 32.55 \textcolor{gray!70}{\scriptsize (-0.3\%)} & 28.84 & 0.67 & 0.40 \\

\textbf{DBCache}
& 16.88 \textcolor{gray!70}{\scriptsize (-27.4\%)} & 1.38$\times$ & 2384.29 & 1.56$\times$ & 1.01 \textcolor{gray!70}{\scriptsize (+2.0\%)} & 32.53 \textcolor{gray!70}{\scriptsize (-0.3\%)} & 33.86 & 0.87 & 0.12 \\

\midrule

\textbf{FORA} ($\mathcal{N}$=3) 
& \textbf{9.06} \textcolor{gray!70}{\scriptsize (-61.0\%)} & \textbf{2.57}$\times$ & \textbf{1267.89} & \textbf{2.94}$\times$ & 0.93 \textcolor{gray!70}{\scriptsize (-6.1\%)} & \textbf{32.89} \textcolor{gray!70}{\scriptsize (+0.8\%)} & 28.86 & 0.66 & 0.40 \\

\textbf{TeaCache} ($l$=0.6) 
& \underline{9.13} \textcolor{gray!70}{\scriptsize (-60.7\%)} & \underline{2.55}$\times$ & \underline{1342.20} & \underline{2.78}$\times$ & 0.91 \textcolor{gray!70}{\scriptsize (-8.1\%)} & 32.11 \textcolor{gray!70}{\scriptsize (-1.6\%)} & 29.03 & 0.68 & 0.40  \\

\textbf{TaylorSeer} ($\mathcal{N}$=3, $O$=2)
& 10.16 \textcolor{gray!70}{\scriptsize (-56.3\%)} & 2.29$\times$ & 1416.92 & 2.63$\times$ & \textbf{1.01} \textcolor{gray!70}{\scriptsize (+2.0\%)} & \underline{32.86} \textcolor{gray!70}{\scriptsize (+0.7\%)} & \underline{30.77} & \underline{0.78} & \underline{0.23}  \\

\rowcolor{gray!20}
\textbf{FreqCa} ($\mathcal{N}$=3) 
& 9.37 \textcolor[HTML]{0f98b0}{\scriptsize (-59.7\%)} & 2.48$\times$ & 1417.40 & 2.63$\times$ & \underline{1.00} \textcolor[HTML]{0f98b0}{\scriptsize (+1.0\%)} & 32.61 \textcolor[HTML]{0f98b0}{\scriptsize (-0.1\%)} & \textbf{33.03} & \textbf{0.86} & \textbf{0.13}  \\

\midrule

\textbf{FORA}($\mathcal{N}$=5) {\textcolor{red}{$^{\dagger}$}}
& \underline{5.97} \textcolor{gray!70}{\scriptsize (-74.3\%)} & \underline{3.89}$\times$ & 820.80 & 4.54$\times$ & 0.82 \textcolor{gray!70}{\scriptsize (-17.2\%)} & 32.48 \textcolor{gray!70}{\scriptsize (-0.5\%)} & 28.44 & 0.60 & 0.50  \\ 

\textbf{\texttt{ToCa}}($\mathcal{N}$=8, $\mathcal{R}$=75\%) 
& 12.39 \textcolor{gray!70}{\scriptsize (-46.7\%)} & 1.88$\times$ & 829.86 & 4.49$\times$ & 0.95 \textcolor{gray!70}{\scriptsize (-4.0\%)} & 32.60 \textcolor{gray!70}{\scriptsize (-0.1\%)} & \underline{29.07} & 0.64 & 0.43   \\ 

\textbf{\texttt{DuCa}}($\mathcal{N}$=8, $\mathcal{R}$=70\%) 
& 9.40 \textcolor{gray!70}{\scriptsize (-59.6\%)} & 2.47$\times$ & 858.27 & 4.34$\times$ & 0.94 \textcolor{gray!70}{\scriptsize (-5.1\%)} & 32.58 \textcolor{gray!70}{\scriptsize (-0.2\%)} & 29.06 & 0.64 & 0.43   \\ 

\textbf{TeaCache}($l$=1.0) {\textcolor{red}{$^{\dagger}$}}
& 7.07 \textcolor{gray!70}{\scriptsize (-69.6\%)} & 3.29$\times$ & 820.55 & 4.54$\times$ & 0.84 \textcolor{gray!70}{\scriptsize (-15.2\%)} & 31.88 \textcolor{gray!70}{\scriptsize (-2.3\%)} & 28.61 & 0.64 & 0.48  \\

\textbf{TaylorSeer}($\mathcal{N}$=6, $O$=2)
& 6.73 \textcolor{gray!70}{\scriptsize (-71.0\%)} & 3.45$\times$ & \underline{746.28} & \textbf{4.99}$\times$ & \underline{1.00} \textcolor{gray!70}{\scriptsize (+1.0\%)} & \underline{32.91} \textcolor{gray!70}{\scriptsize (+0.8\%)} & 28.94 & \underline{0.66} & \underline{0.40}  \\

\rowcolor{gray!20}
\textbf{FreqCa} ($\mathcal{N}$=7)
& \textbf{5.19} \textcolor[HTML]{0f98b0}{\scriptsize (-77.7\%)} & \textbf{4.48}$\times$ & \textbf{746.03 } & \underline{4.99}$\times$ & \textbf{1.01} \textcolor[HTML]{0f98b0}{\scriptsize (+2.0\%)} & \textbf{32.98} \textcolor[HTML]{0f98b0}{\scriptsize (+1.0\%)} & \textbf{30.00} & \textbf{0.71} & \textbf{0.31} \\

\midrule

\textbf{FORA}($\mathcal{N}$=7) {\textcolor{red}{$^{\dagger}$}}
& \underline{5.09} \textcolor{gray!70}{\scriptsize (-78.1\%)} & \underline{4.57}$\times$ & \textbf{597.25} & \textbf{6.24}$\times$ & 0.68 \textcolor{gray!70}{\scriptsize (-31.3\%)} & 31.90 \textcolor{gray!70}{\scriptsize (-2.3\%)} & 28.32 & 0.59 & 0.54  \\ 

\textbf{\texttt{ToCa}}($\mathcal{N}$=12, $\mathcal{R}$=85\%) {\textcolor{red}{$^{\dagger}$}}
& 9.82 \textcolor{gray!70}{\scriptsize (-57.7\%)} & 2.37$\times$ & 618.57 & 6.02$\times$ & 0.80 \textcolor{gray!70}{\scriptsize (-19.2\%)} & \underline{32.32} \textcolor{gray!70}{\scriptsize (-1.0\%)} & 28.70 & 0.60 & 0.52   \\ 

\textbf{\texttt{DuCa}}($\mathcal{N}$=12, $\mathcal{R}$=80\%) {\textcolor{red}{$^{\dagger}$}}
& 7.74 \textcolor{gray!70}{\scriptsize (-66.7\%)} & 3.00$\times$ & 646.97 & 5.76$\times$ & 0.77 \textcolor{gray!70}{\scriptsize (-22.2\%)} & 32.20 \textcolor{gray!70}{\scriptsize (-1.3\%)} & \underline{28.71} & \underline{0.60} & 0.53   \\ 

\textbf{TeaCache}($l$=1.4) {\textcolor{red}{$^{\dagger}$}}
& 6.14 \textcolor{gray!70}{\scriptsize (-73.6\%)} & 3.79$\times$ & 671.51 & 5.55$\times$ & 0.74 \textcolor{gray!70}{\scriptsize (-25.3\%)} & 31.78 \textcolor{gray!70}{\scriptsize (-2.6\%)} & 28.12 & 0.48 & 0.68  \\

\textbf{TaylorSeer}($\mathcal{N}$=9, $O$=2) {\textcolor{red}{$^{\dagger}$}}
& 5.85 \textcolor{gray!70}{\scriptsize (-74.8\%)} & 3.97$\times$ & \textbf{597.25} & \textbf{6.24}$\times$ & \underline{0.86} \textcolor{gray!70}{\scriptsize (-13.1\%)} & 32.04 \textcolor{gray!70}{\scriptsize (-1.8\%)} & 28.38 & 0.59 & \underline{0.51}  \\

\rowcolor{gray!20}
\textbf{FreqCa} ($\mathcal{N}$=10)
& \textbf{4.25} \textcolor[HTML]{0f98b0}{\scriptsize (-81.7\%)} & \textbf{5.47}$\times$ & \underline{597.47} & \underline{6.24}$\times$ & \textbf{0.97} \textcolor[HTML]{0f98b0}{\scriptsize (-2.0\%)} & \textbf{32.53} \textcolor[HTML]{0f98b0}{\scriptsize (-0.3\%)} & \textbf{28.77} & \textbf{0.62} & \textbf{0.43}  \\
\midrule
\textbf{[schnell]: 4 steps} 
& 2.17\textcolor{gray!70}{\scriptsize (+0.0\%)} & 1.00$\times$ & 278.41 & 1.00$\times$ & 0.93\textcolor{gray!70}{\scriptsize (+0.0\%)} & 34.09\textcolor{gray!70}{\scriptsize (+0.0\%)} & $\infty$ & 1.00 & 0.00 \\

\rowcolor{gray!20}
\textbf{FreqCa ($\mathcal{N}$=3): 4 steps}
& 1.29 \textcolor[HTML]{0f98b0}{\scriptsize (-40.6\%)} 
& 1.68$\times$ 
& 139.23 
& 2.00$\times$ 
& 0.95 \textcolor[HTML]{0f98b0}{\scriptsize (+2.2\%)} 
& 34.47 \textcolor[HTML]{0f98b0}{\scriptsize (+1.1\%)} 
& 30.30 
& 0.81 
& 0.16 \\
\bottomrule
\end{tabular}
}

\label{table:FLUX-Metrics}
\raggedright
{\scriptsize
\begin{itemize}[leftmargin=10pt,topsep=0pt]
\item\textcolor{red}{$\dagger$} Methods exhibit significant degradation in image quality. 
\vspace{-1.5mm}
\item\textcolor{gray!70}{Gray}: Baseline-relative degradation in quality and gains in latency. \textcolor[HTML]{0f98b0}{Blue}: \textbf{FreqCa} achieves minimal degradation with large latency gains.
\end{itemize}
}
\vspace{-2mm}
\end{table*}

On FLUX.1-dev, \textit{FreqCa} consistently outperforms state-of-the-art acceleration methods across different speedup levels. At \textbf{2.63$\times$} speedup, \textit{FreqCa} achieves an \textbf{ImageReward of 1.00}, clearly outperforming FORA and TeaCache. At \textbf{4.99$\times$} speedup, it maintains lossless quality. Even under \textbf{6.24$\times$} speedup, \textit{FreqCa} achieves only a \textbf{2\%} drop in ImageReward (0.97), while TaylorSeer suffers a degradation of \textbf{13.1\%}. \textit{FreqCa} also achieves \textbf{2.00$\times$} speedup on distilled FLUX.1-schnell while improving ImageReward from 0.93 to 0.95.

\vspace{-1.5mm}
\subsubsection{Qwen-Image}
\vspace{-1.5mm}
\begin{table*}[htb]
\centering
\caption{\textbf{Quantitative comparison in text-to-image generation} for Qwen-Image and Qwen-Image-Lightning (a distilled version). Best results are highlighted in \textbf{bold}, and second-best are \underline{underlined}.}
\vspace{-3mm}
  \resizebox{\textwidth}{!}{ 
    \begin{tabular}{l | c  c | c  c | c  c | c c c}
        \toprule
        \multirow{2}{*}{\textbf{Method}}
        & \multicolumn{4}{c|}{\textbf{Acceleration}} 
        & \multicolumn{2}{c|}{\textbf{Quality Metrics}} 
        & \multicolumn{3}{c}{\textbf{Perceptual Metrics}}\rule{0pt}{2ex}\\
        \cline{2-10}
        & \textbf{Latency(s) \(\downarrow\)} 
        & \textbf{Speed \(\uparrow\)} 
        & \textbf{FLOPs(T) \(\downarrow\)}  
        & \textbf{Speed \(\uparrow\)} 
        & \textbf{ImageReward\(\uparrow\)} 
        & \textbf{CLIP\(\uparrow\)} 
        & \textbf{PSNR\(\uparrow\)} 
        & \textbf{SSIM\(\uparrow\)}
        & \textbf{LPIPS\(\downarrow\)}\rule{0pt}{2ex}\\
        \midrule

\textbf{$50$ steps}
& 127.40 \textcolor{gray!70}{\scriptsize (+0.0\%)} & 1.00$\times$ & 12917.56 & 1.00$\times$ & 1.25 \textcolor{gray!70}{\scriptsize (+0.0\%)} & 35.59 \textcolor{gray!70}{\scriptsize (+0.0\%)} & $\infty$ & 1.00 & 0.00 \\

\textbf{$50\%$ steps }
& 64.10 \textcolor{gray!70}{\scriptsize (-49.6\%)} & 1.99$\times$ & 6458.78 & 2.00$\times$ & 1.20 \textcolor{gray!70}{\scriptsize (-4.0\%)} & 35.31 \textcolor{gray!70}{\scriptsize (-0.6\%)} & 30.54 & 0.75 & 0.28 \\

\textbf{$20\%$ steps} {\textcolor{red}{$^{\dagger}$}}
& 25.92 \textcolor{gray!70}{\scriptsize (-79.6\%)} & 4.89$\times$ & 2583.51 & 5.00$\times$ & 0.94 \textcolor{gray!70}{\scriptsize (-26.4\%)} & 34.95 \textcolor{gray!70}{\scriptsize (-1.0\%)} & 28.59 & 0.61 & 0.52 \\

\midrule

\textbf{FORA}($\mathcal{N}$=4){\textcolor{red}{$^{\dagger}$}}
& 38.43 \textcolor{gray!70}{\scriptsize (-70.0\%)} & 3.32$\times$ & 3359.99 & 3.84$\times$ & 0.93 \textcolor{gray!70}{\scriptsize (-25.6\%)} & 34.40 \textcolor{gray!70}{\scriptsize (-3.3\%)} & 28.66 & 0.59 & 0.51 \\

\textbf{\texttt{ToCa}}($\mathcal{N}$=8, $\mathcal{R}$=75\%){\textcolor{red}{$^{\dagger}$}}
& 61.37 \textcolor{gray!70}{\scriptsize (-51.8\%)} & 2.08$\times$ & 2991.34 & 4.32$\times$ & \underline{1.02} \textcolor{gray!70}{\scriptsize (-18.4\%)} & \underline{34.96} \textcolor{gray!70}{\scriptsize (-1.8\%)} & \underline{28.93} & \underline{0.63} & \underline{0.44} \\

\textbf{\texttt{DuCa}}($\mathcal{N}$=9, $\mathcal{R}$=80\%){\textcolor{red}{$^{\dagger}$}}
& 34.73 \textcolor{gray!70}{\scriptsize (-72.7\%)} & 3.67$\times$ & 2958.13 & 4.37$\times$ & 0.77 \textcolor{gray!70}{\scriptsize (-38.4\%)} & 34.62 \textcolor{gray!70}{\scriptsize (-2.7\%)} & 28.45 & 0.58 & 0.55 \\

\textbf{TaylorSeer}($\mathcal{N}$=6)
& \underline{30.75} \textcolor{gray!70}{\scriptsize (-75.9\%)} & \underline{4.14}$\times$ & \textbf{2583.97} & \textbf{5.00}$\times$ & 1.01 \textcolor{gray!70}{\scriptsize (-19.2\%)} & 34.71 \textcolor{gray!70}{\scriptsize (-2.5\%)} & 28.58 & 0.62 & 0.46 \\

\rowcolor{gray!20}
\textbf{FreqCa}($\mathcal{N}$=6)
& \textbf{29.80} \textcolor[HTML]{0f98b0}{\scriptsize (-76.6\%)} & \textbf{4.28}$\times$ & \underline{2584.70} & \underline{5.00}$\times$ & \textbf{1.20} \textcolor[HTML]{0f98b0}{\scriptsize (-4.0\%)} & \textbf{35.39} \textcolor[HTML]{0f98b0}{\scriptsize (-0.6\%)} & \textbf{29.67} & \textbf{0.78} & \textbf{0.33} \\

\midrule

\textbf{FORA}($\mathcal{N}$=6){\textcolor{red}{$^{\dagger}$}}
& 28.69 \textcolor{gray!70}{\scriptsize (-77.5\%)} & 4.44$\times$ & 2326.74 & 5.55$\times$ & 0.48 \textcolor{gray!70}{\scriptsize (-61.6\%)} & 33.34 \textcolor{gray!70}{\scriptsize (-6.3\%)} & 28.48 & 0.55 & 0.59 \\

\textbf{\texttt{ToCa}}($\mathcal{N}$=12, $\mathcal{R}$=85\%){\textcolor{red}{$^{\dagger}$}}
& 50.95 \textcolor{gray!70}{\scriptsize (-60.0\%)} & 2.50$\times$ & 2406.20 & 5.37$\times$ & 0.55 \textcolor{gray!70}{\scriptsize (-56.0\%)} & \underline{34.08} \textcolor{gray!70}{\scriptsize (-4.2\%)} & \underline{28.69} & \underline{0.57} & \underline{0.53} \\

\textbf{\texttt{DuCa}}($\mathcal{N}$=12, $\mathcal{R}$=90\%){\textcolor{red}{$^{\dagger}$}}
& 28.57 \textcolor{gray!70}{\scriptsize (-77.6\%)} & 4.46$\times$ & 2171.56 & 5.95$\times$ & 0.41 \textcolor{gray!70}{\scriptsize (-67.2\%)} & 33.38 \textcolor{gray!70}{\scriptsize (-6.2\%)} & 28.38 & \underline{0.57} & 0.60 \\

\textbf{TaylorSeer}($\mathcal{N}$=9){\textcolor{red}{$^{\dagger}$}}
& \underline{24.64} \textcolor{gray!70}{\scriptsize (-80.7\%)} & \underline{5.17}$\times$ & \underline{2067.29} & \underline{6.25}$\times$ & \underline{0.73} \textcolor{gray!70}{\scriptsize (-41.6\%)} & 32.97 \textcolor{gray!70}{\scriptsize (-7.4\%)} & 28.25 & 0.56 & 0.58 \\

\rowcolor{gray!20}
\textbf{FreqCa}($\mathcal{N}$=10)
& \textbf{22.45} \textcolor[HTML]{0f98b0}{\scriptsize (-82.4\%)} & \textbf{5.68}$\times$ & \textbf{1809.38} & \textbf{7.14}$\times$ & \textbf{1.02} \textcolor[HTML]{0f98b0}{\scriptsize (-18.4\%)} & \textbf{35.00} \textcolor[HTML]{0f98b0}{\scriptsize (-1.7\%)} & \textbf{28.86} & \textbf{0.64} & \textbf{0.44} \\

\midrule
\textbf{Qwen-Image-Lightning: 8 steps} 
& 7.27\textcolor{gray!70}{\scriptsize (+0.0\%)}  & 1.00$\times$ & 560.96 & 1.00$\times$ & 1.30\textcolor{gray!70}{\scriptsize (+0.0\%)} & 35.26\textcolor{gray!70}{\scriptsize (+0.0\%)} & $\infty$ & 1.00 & 0.00 \\

\rowcolor{gray!20}
\textbf{FreqCa ($\mathcal{N}$=2): 8 steps}
& 5.29 \textcolor[HTML]{0f98b0}{\scriptsize (-27.2\%)} 
& 1.37$\times$ 
& 350.61 
& 1.60$\times$ 
& \textbf{1.29} \textcolor[HTML]{0f98b0}{\scriptsize (-0.8\%)} 
& 35.23 \textcolor[HTML]{0f98b0}{\scriptsize (-0.1\%)} 
& \textbf{33.01} 
& \textbf{0.83} 
& \textbf{0.12} \\

\rowcolor{gray!20}
\textbf{FreqCa ($\mathcal{N}$=3): 8 steps}
& 4.63 \textcolor[HTML]{0f98b0}{\scriptsize (-36.3\%)} 
& 1.57$\times$ 
& 282.26 
& 1.99$\times$ 
& 1.28 \textcolor[HTML]{0f98b0}{\scriptsize (-1.5\%)} 
& 35.10 \textcolor[HTML]{0f98b0}{\scriptsize (-0.5\%)} 
& 31.36 
& 0.77 
& 0.18 \\

\rowcolor{gray!20}
\textbf{FreqCa ($\mathcal{N}$=4): 8 steps}
& \textbf{3.29} \textcolor[HTML]{0f98b0}{\scriptsize (-54.9\%)} 
& \textbf{2.21}$\times$ 
& \textbf{140.25} 
& \textbf{4.00}$\times$ 
& 1.29 \textcolor[HTML]{0f98b0}{\scriptsize (-0.8\%)} 
& \textbf{35.94} \textcolor[HTML]{0f98b0}{\scriptsize (+1.9\%)} 
& 29.25 
& 0.62
& 0.36 \\
\bottomrule
\end{tabular}
}

\label{table:qwen-image-Metrics}
\raggedright
{\scriptsize
\begin{itemize}[leftmargin=10pt,topsep=0pt]
\item\textcolor{red}{$\dagger$} Methods exhibit significant degradation in image quality. 
\vspace{-1.5mm}
\item\textcolor{gray!70}{Gray}: Baseline-relative degradation in quality and gains in latency. \textcolor[HTML]{0f98b0}{Blue}: \textbf{FreqCa} achieves minimal degradation with large latency gains.
\end{itemize}
}
\vspace{-2mm}
\end{table*}

On Qwen-Image, \textit{FreqCa} demonstrates superior performance across different acceleration levels. At \textbf{5.00$\times$} speedup, \textit{FreqCa} achieves an ImageReward of \textbf{1.20}, outperforming TaylorSeer (1.01). At \textbf{7.14$\times$} speedup, \textit{FreqCa} shows only a \textbf{18.4\%} drop in ImageReward (1.02), while TaylorSeer suffers a \textbf{41.6\%} quality loss (0.73). \textit{FreqCa} achieves \textbf{4.00$\times$} speedup on distilled Qwen-Image-Lightning with minimal quality degradation.

\vspace{-1.5mm}
\subsection{Image Editing}
\vspace{-1.5mm}
\subsubsection{FLUX.1-Kontext-dev}
\vspace{-1.5mm}

On FLUX.1-Kontext-dev, \textit{FreqCa} outperforms other acceleration methods. At \textbf{5.00$\times$} speedup, \textit{FreqCa} achieves a Q\_O score of \textbf{6.195}, outperforming ToCa (6.125). At \textbf{6.24$\times$} speedup, \textit{FreqCa} shows only a \textbf{0.4\%} drop in Q\_O score, demonstrating better perceptual fidelity.
\begin{table*}[hbt]
\centering
\caption{\textbf{Quantitative comparison of text-to-image generation} for 
FLUX.1-Kontext-dev. Best results are highlighted
in \textbf{bold}, and second-best results are \underline{underlined}.}
\vspace{-3mm}
\resizebox{\textwidth}{!}{ 
    \begin{tabular}{l | c  c | c  c | c c c}
    \toprule
    \multirow[c]{2}{*}{\bf Method}
    & \multicolumn{4}{c|}{\bf Acceleration} 

    & \multicolumn{3}{c}{\bf GEdit-EN (Full)} \rule{0pt}{2ex}\\
    \cline{2-8}
    & {\bf Latency(s) $\downarrow$} 
    & {\bf Speed $\uparrow$} 
    & {\bf FLOPs(T) $\downarrow$}  
    & {\bf Speed $\uparrow$} 
    
    & {\bf Q\_SC $\uparrow$} 
    & {\bf Q\_PQ $\uparrow$} 
    & {\bf Q\_O $\uparrow$}\rule{0pt}{2ex}\\
    \midrule

\textbf{[Kontext]50 steps}
& 50.20 \textcolor{gray!70}{\scriptsize (+0.0\%)} & 1.00$\times$ & 8299.54 & 1.00$\times$ & 6.481 & 7.331 & 6.213 \textcolor{gray!70}{\scriptsize (+0.0\%)} \\

\textbf{50\% steps} {\textcolor{red}{$^{\dagger}$}}
& 25.42 \textcolor{gray!70}{\scriptsize (-49.4\%)} & 1.97$\times$ & 4149.77 & 2.00$\times$ & 6.544 & 7.286 & 6.253 \textcolor{gray!70}{\scriptsize (+0.6\%)} \\
\textbf{20\% steps} {\textcolor{red}{$^{\dagger}$}}
& 10.47 \textcolor{gray!70}{\scriptsize (-79.1\%)} & 4.79$\times$ & 1659.91 & 5.00$\times$ & 6.603 & 7.184 & 6.286 \textcolor{gray!70}{\scriptsize (+1.2\%)} \\

\midrule

\textbf{\texttt{ToCa}}($\mathcal{N}$=8, $\mathcal{R}$=70\%) 
& 29.56 \textcolor{gray!70}{\scriptsize (-41.1\%)} & 1.70$\times$ & 1841.35 & 4.51$\times$ & 6.432 & 7.256 & 6.125 \textcolor{gray!70}{\scriptsize (-1.4\%)} \\

\textbf{\texttt{DuCa}}($\mathcal{N}$=8, $\mathcal{R}$=60\%) 
& 13.12 \textcolor{gray!70}{\scriptsize (-73.9\%)} & 3.83$\times$ & 1669.08 & 4.97$\times$ & 6.469 & 7.195 & 6.150 \textcolor{gray!70}{\scriptsize (-1.0\%)} \\


\textbf{TaylorSeer}($\mathcal{N}$=6, $\mathcal{O}$=2)
& \underline{13.95} \textcolor{gray!70}{\scriptsize (-72.2\%)} & \underline{3.60}$\times$ & \textbf{1660.95} & \underline{5.00}$\times$ & \underline{6.477} & \textbf{7.296} & \underline{6.170} \textcolor{gray!70}{\scriptsize (-0.7\%)} \\

\rowcolor{gray!20}
\textbf{FreqCa}($\mathcal{N}$=7)
& \textbf{10.77} \textcolor[HTML]{0f98b0}{\scriptsize (-78.5\%)} & \textbf{4.66}$\times$ & \underline{1661.37} & \textbf{5.00}$\times$ & \textbf{6.480} & \underline{7.271} & \textbf{6.195} \textcolor[HTML]{0f98b0}{\scriptsize (-0.3\%)} \\

\midrule


\textbf{\texttt{ToCa}}($\mathcal{N}$=12, $\mathcal{R}$=75\%)
& 20.72 \textcolor{gray!70}{\scriptsize (-58.7\%)} & 2.42$\times$ & 1359.61 & 6.10$\times$ & 6.393 & 6.919 & 6.041 \textcolor{gray!70}{\scriptsize (-2.8\%)}  \\

\textbf{\texttt{DuCa}}($\mathcal{N}$=12, $\mathcal{R}$=70\%) 
& 10.39 \textcolor{gray!70}{\scriptsize (-79.3\%)} & 4.83$\times$ & 1376.55 & 6.03$\times$ & 6.597 & 7.005 & 6.173 \textcolor{gray!70}{\scriptsize (-0.6\%)} \\


\textbf{TaylorSeer}($\mathcal{N}$=9, $\mathcal{O}$=2)
& \underline{12.05} \textcolor{gray!70}{\scriptsize (-76.0\%)} & \underline{4.17}$\times$ & \textbf{1329.02} & \underline{6.24}$\times$ & \underline{6.407} & \underline{6.995} & \underline{6.074} \textcolor{gray!70}{\scriptsize (-2.2\%)} \\

\rowcolor{gray!20}
\textbf{FreqCa}($\mathcal{N}$=10)
& \textbf{8.74} \textcolor[HTML]{0f98b0}{\scriptsize (-82.6\%)} & \textbf{5.74}$\times$ & \underline{1329.33} & \textbf{6.24}$\times$ & \textbf{6.550} & \textbf{7.160} & \textbf{6.190} \textcolor[HTML]{0f98b0}{\scriptsize (-0.4\%)} \\

\bottomrule
\end{tabular}
}

\label{table:kontext-Metrics}
\raggedright
{\scriptsize
\begin{itemize}[leftmargin=10pt,topsep=0pt]
\item\textcolor{red}{$\dagger$} Methods exhibit significant degradation in image quality;  \textbf{Q\_SC}: semantic consistency,\textbf{ Q\_PQ}: perceptual quality, \textbf{Q\_O}: overall score. 
\vspace{-1.5mm}
\item\textcolor{gray!70}{Gray}: Baseline-relative degradation in quality and gains in latency. \textcolor[HTML]{0f98b0}{Blue}: \textbf{FreqCa} achieves minimal degradation with large latency gains.
\end{itemize}
}
\vspace{-2mm}
\end{table*}

\subsubsection{Qwen-Image-Edit}
\vspace{-3mm}

\begin{figure}[htbp]
    \centering
    \includegraphics[width=\linewidth]
    {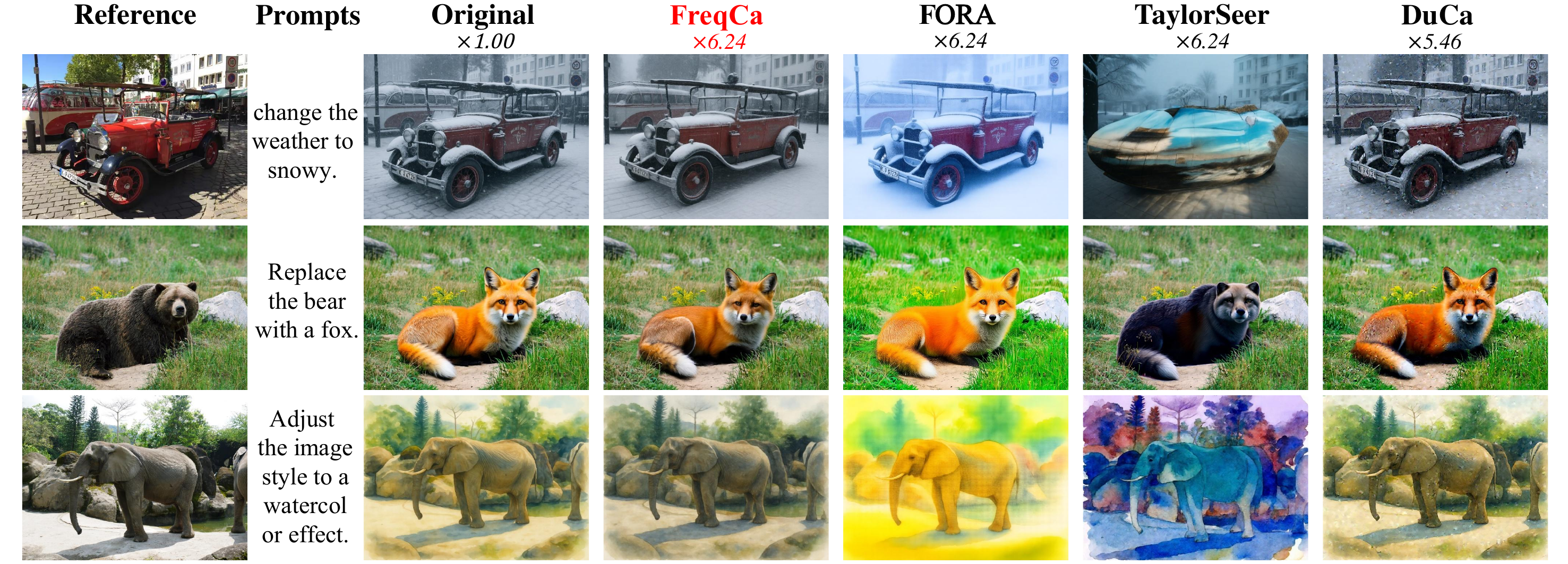}
    \captionof{figure}{On 
    Qwen-Image-Edit, \textit{FreqCa} delivers higher speedup with near-original editing quality }

    \label{fig:qwen-image-edit}
    \vspace{-2mm}
\end{figure}

\begin{table*}[htb]
\centering
\caption{\textbf{Quantitative comparison of image editing} using Qwen-Image-Edit. Best results are highlighted
in \textbf{bold}, and second-best results are \underline{underlined}.}
\vspace{-3mm}
  \resizebox{\textwidth}{!}{ 
    \begin{tabular}{l | c  c | c  c | c c c|c c c}
    \toprule
    \multirow[c]{2}{*}{\bf Method}
    & \multicolumn{4}{c|}{\bf Acceleration} 
    & \multicolumn{3}{c|}{\bf GEdit-CN (Full)}
    & \multicolumn{3}{c}{\bf GEdit-EN (Full)} \rule{0pt}{2ex}\\
    \cline{2-11}
    & {\bf Latency(s) $\downarrow$} 
    & {\bf Speed $\uparrow$} 
    & {\bf FLOPs(T) $\downarrow$}  
    & {\bf Speed $\uparrow$} 
    & {\bf Q\_SC $\uparrow$} 
    & {\bf Q\_PQ $\uparrow$} 
    & {\bf Q\_O $\uparrow$}
    & {\bf Q\_SC $\uparrow$} 
    & {\bf Q\_PQ $\uparrow$} 
    & {\bf Q\_O $\uparrow$}\rule{0pt}{2ex}\\
    \midrule

\textbf{[full]50 steps}
& 284.51 \textcolor{gray!70}{\scriptsize (+0.0\%)} & 1.00$\times$ & 28190.88 & 1.00$\times$ & 7.68 & 7.51 & 7.41 \textcolor{gray!70}{\scriptsize (+0.0\%)} & 7.82 & 7.54 & 7.54 \textcolor{gray!70}{\scriptsize (+0.0\%)}\\

\textbf{50\% steps} {\textcolor{red}{$^{\dagger}$}}
& 143.29 \textcolor{gray!70}{\scriptsize (-49.6\%)} & 1.99$\times$ & 14095.44 & 2.00$\times$ & 7.70 & 7.53 & 7.44 \textcolor{gray!70}{\scriptsize (+0.4\%)} & 7.77 & 7.52 & 7.47 \textcolor{gray!70}{\scriptsize (-0.9\%)}\\

\textbf{20\% steps} {\textcolor{red}{$^{\dagger}$}}
& 58.45 \textcolor{gray!70}{\scriptsize (-79.5\%)} & 4.87$\times$ & 5638.18 & 5.00$\times$ & 7.65 & 7.42 & 7.35 \textcolor{gray!70}{\scriptsize (-0.8\%)} & 7.73 & 7.46 & 7.44 \textcolor{gray!70}{\scriptsize (-1.3\%)}\\

\midrule
\textbf{FORA}($\mathcal{N}$=5)
& \underline{63.15} \textcolor{gray!70}{\scriptsize (-77.8\%)} & \underline{4.51}$\times$ & \underline{5643.13} & \underline{5.00}$\times$ & 7.60 & 7.31 & 7.25\textcolor{gray!70}{\scriptsize (-2.2\%)} & 7.62 & 7.34 & 7.28 \textcolor{gray!70}{\scriptsize (-3.4\%)}\\

\textbf{\texttt{DuCa}}($\mathcal{N}$=7, $\mathcal{R}$=95\%) & 69.54 \textcolor{gray!70}{\scriptsize (-75.5\%)} & 4.09$\times$ & 5699.89 & 4.95$\times$ & \underline{7.73} & \underline{7.44} & \underline{7.44} \textcolor{gray!70}{\scriptsize (+0.4\%)} & \underline{7.80} & \underline{7.40}& \underline{7.45} \textcolor{gray!70}{\scriptsize (-1.2\%)} \\

\textbf{TaylorSeer}($\mathcal{N}$=6, $\mathcal{O}$=2)
&  65.66 \textcolor{gray!70}{\scriptsize (-76.9\%)} & 4.33$\times$ & \underline{5643.13} & \underline{5.00}$\times$ & 7.25 & 7.09 & 6.92 \textcolor{gray!70}{\scriptsize (-6.6\%)} & 7.26 & 7.14 & 6.89 \textcolor{gray!70}{\scriptsize (-8.6\%)}\\

\rowcolor{gray!20}
\textbf{FreqCa} ($\mathcal{N}$=6)
& \textbf{62.89} \textcolor[HTML]{0f98b0}{\scriptsize (-77.9\%)} & \textbf{4.52}$\times$ & \textbf{5642.24} & \textbf{5.00}$\times$ & \textbf{7.75} & \textbf{7.54} & \textbf{7.49} \textcolor[HTML]{0f98b0}{\scriptsize (+1.1\%)} & \textbf{7.83} & \textbf{7.48} & \textbf{7.52} \textcolor[HTML]{0f98b0}{\scriptsize (-0.3\%)}\\

\midrule

\textbf{FORA}($\mathcal{N}$=7)
& \underline{52.20} \textcolor{gray!70}{\scriptsize (-81.7\%)} & \underline{5.45}$\times$ & \underline{4515.74} & \underline{6.24}$\times$ & 7.42 & \underline{7.13} & \underline{7.06} \textcolor{gray!70}{\scriptsize (-4.7\%)} & 7.43 & \textbf{7.19} & \underline{7.06} \textcolor{gray!70}{\scriptsize (-6.3\%)}\\

\textbf{\texttt{DuCa}}($\mathcal{N}$=10, $\mathcal{R}$=95\%)& 59.81 \textcolor{gray!70}{\scriptsize (-79.0\%)} & 4.76$\times$ & 5158.45 & 5.46$\times$ & \underline{7.50} & 5.75 & 6.39 \textcolor{gray!70}{\scriptsize (-13.8\%)} & \underline{7.52} & 5.77 & 6.41 \textcolor{gray!70}{\scriptsize (-15.0\%)} \\

\textbf{TaylorSeer}($\mathcal{N}$=9, $\mathcal{O}$=2)
& 53.92 \textcolor{gray!70}{\scriptsize (-81.1\%)} & 5.28$\times$ & \underline{4515.74}& \underline{6.24}$\times$ & 6.61 & 6.65 & 6.31 \textcolor{gray!70}{\scriptsize (-14.8\%)} & 6.67 & 6.63 & 6.31 \textcolor{gray!70}{\scriptsize (-16.3\%)}\\

\rowcolor{gray!20}
\textbf{FreqCa }($\mathcal{N}$=9)
& \textbf{51.09} \textcolor[HTML]{0f98b0}{\scriptsize (-82.0\%)} & \textbf{5.57}$\times$ & \textbf{4514.48} & \textbf{6.24}$\times$& \textbf{7.62} & \textbf{7.18} & \textbf{7.27} \textcolor[HTML]{0f98b0}{\scriptsize (-1.9\%)} & \textbf{7.66} & \underline{7.12} & \textbf{7.21} \textcolor[HTML]{0f98b0}{\scriptsize (-4.3\%)}\\

\bottomrule
\end{tabular}
}
\label{table:qwen-edit-Metrics}
\raggedright
{\scriptsize
\begin{itemize}[leftmargin=10pt,topsep=0pt]
\item\textcolor{red}{$\dagger$} Methods exhibit significant degradation in image quality;  \textbf{Q\_SC}: semantic consistency,\textbf{ Q\_PQ}: perceptual quality, \textbf{Q\_O}: overall score. 
\vspace{-1.5mm}
\item\textcolor{gray!70}{Gray}: Baseline-relative degradation in quality and gains in latency. \textcolor[HTML]{0f98b0}{Blue}: \textbf{FreqCa} achieves minimal degradation with large latency gains.
\end{itemize}
}
\vspace{-2mm}
\end{table*}

On Qwen-Image-Edit, \textit{FreqCa} demonstrates superior performance in bilingual editing tasks. At \textbf{5.00$\times$} speedup, \textit{FreqCa} achieves Q\_O scores of {7.49} on GEdit-CN and {7.52} on GEdit-EN, outperforming TaylorSeer (6.92 and 6.89). At \textbf{6.24$\times$} speedup, \textit{FreqCa} shows quality drops of only \textbf{1.9\%} and \textbf{4.3\%}, while TaylorSeer degrades by {14.8\%} and {16.3\%}.


As shown in \textbf{Figures \ref{fig:quality} and \ref{fig:qwen-image-edit} }, qualitative evaluation confirms \textit{FreqCa}'s superior visual quality preservation. While FORA (6.24$\times$), Duca(5.46$\times$) and TaylorSeer (6.24$\times$) exhibit significant artifacts, \textit{FreqCa} (6.24$\times$) maintains consistent visual quality comparable to the original model.

\vspace{-1.5mm}

\subsection{Ablation Studies}
\vspace{-1.5mm}

\subsubsection{Cache Memory and Computational Efficiency}
\vspace{-1.5mm}

Conventional layer-wise caching methods store both attention and MLP outputs per layer ($N=2$) and retain $m+1$ historical states for $m$-th order prediction, yielding memory cost $\mathcal{K}_{\text{layer}} = 2(m+1)L$. For FLUX.1-dev ($L=57$) with second-order prediction ($m=2$), this requires 342 cache units.

In contrast, \textit{FreqCa} caches only the CRF, adopting a frequency-decoupled strategy: low-frequency components are reused (1 unit), while high-frequency components employ second-order Hermite interpolation (3 units). The total cost is constant:
\[
\mathcal{K}_{\text{FreqCa}} =1+3= 4, \quad R = \frac{\mathcal{K}_{\text{FreqCa}}}{\mathcal{K}_{\text{layer}}} = \frac{4}{(m+1)\cdot N \cdot L} \approx 1.17\% \quad (m=2,\ L=57,\,N=2),
\]
reducing memory complexity from $\mathcal{O}(L)$ to $\mathcal{O}(1)$. Computationally, prediction steps  incur negligible cost $C_{\text{pred}} \ll C_{\text{full}}$. Executing a full forward pass every $S$ steps yields average cost:
\[
\bar{\mathcal{C}} = \tfrac{1}{S} C_{\text{full}} + (1-\tfrac{1}{S}) C_{\text{pred}} \quad \Rightarrow \quad \text{Speedup} \approx S \quad \text{as } C_{\text{pred}} \to 0.
\]

\textit{FreqCa} achieves near-S$\times$ acceleration with only \textbf{~1\%} additional memory overhead, establishing the first constant-memory, high-throughput inference acceleration framework for diffusion models. 

\begin{figure}[htp]
    \centering
    \begin{minipage}[t]{0.63\linewidth}
        \centering
        \includegraphics[width=\linewidth]{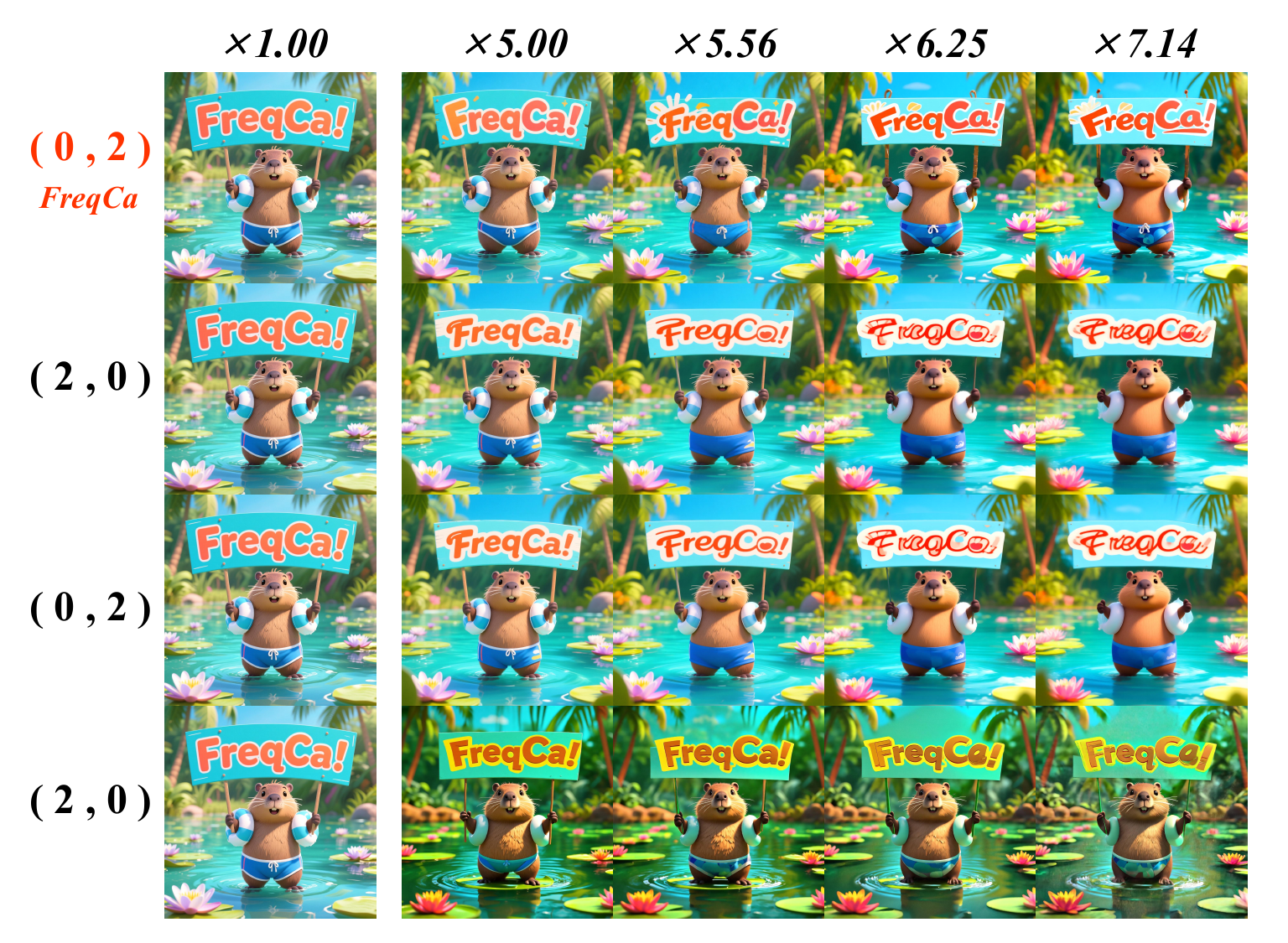}
         \vspace{-4mm}
        \caption{QwenImage ablation results showing image quality under different frequency prediction configurations and acceleration ratios. $(x,y) = (low, high)$ prediction orders.}
        \label{fig:qwen-image-ablation}
        \vspace{-2mm}
    \end{minipage}%
    \hfill
    \begin{minipage}[t]{0.35\linewidth}
        \centering
        \includegraphics[width=\linewidth]{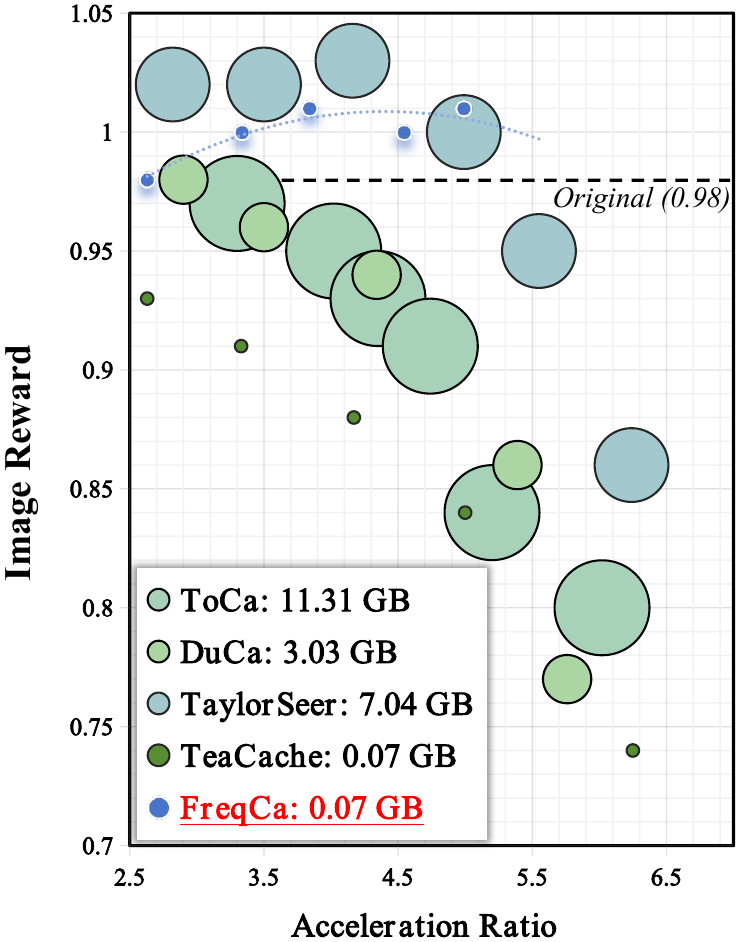}
        \vspace{-4mm}
        \caption{Imagereward versus speedup ratio across methods. Bubble size indicates cache memory.}
    \label{fig:Bubble}
    \vspace{-2mm}
\end{minipage}
\end{figure}

\begin{table}[htb]
\centering
\caption{\textbf{Comparison of methods in Cache Memory, MACs, Latency, and FLOPS on FLUX-1.dev}, Best results are highlighted in \textbf{bold}, and second-best results are \underline{underlined}.}
\vspace{-3mm}
\resizebox{\textwidth}{!}{
\begin{tabular}{l|c|c|c|c|c}
\toprule
\textbf{Method} & \makecell{\bf VRAM\\ \bf Overhead (GB)}$\downarrow$ & \textbf{MACs (T)\(\downarrow\)} & \textbf{Latency (s)\(\downarrow\)} & \textbf{FLOPs (T)\(\downarrow\)} &   \makecell{\bf Image\\ \bf Reward}$\uparrow$\\
\midrule
\textbf{[dev]: 50 steps }         
& 0.62 & 1859.62 & 23.24 & 3726.87 & 0.99 \textcolor{gray!70}{\scriptsize (+0.0\%)}\\
\midrule
\textbf{\texttt{ToCa}}($\mathcal{N}$=8, $\mathcal{R}$=75\%)          
& 12.31\textcolor{gray!70}{\scriptsize (+11.69GB)} & 414.88 & 12.39 & 829.86 & 0.95 \textcolor{gray!70}{\scriptsize (-4.0\%)}\\
\textbf{\texttt{DuCa}}($\mathcal{N}$=8, $\mathcal{R}$=70\%)          
& \underline{3.65}\textcolor{gray!70}{\scriptsize (+3.63GB)} & 428.86 & 9.40 & 858.27 & 0.94 \textcolor{gray!70}{\scriptsize (-5.1\%)}\\
\textbf{TeaCache}($l$=1.0)              
& \textbf{0.69}\textcolor{gray!70}{\scriptsize (+0.07GB)} & 409.43 & 7.07 & 820.55 & 0.84 \textcolor{gray!70}{\scriptsize (-15.2\%)}\\
\textbf{TaylorSeer}($\mathcal{N}$=6, $O$=2)    
& 7.66\textcolor{gray!70}{\scriptsize (+7.02GB)} & \underline{372.38} & \underline{6.73} & \underline{746.28} & \underline{1.00} \textcolor{gray!70}{\scriptsize (+1.0\%)}\\

\textbf{FreqCa}($\mathcal{N}$=7)        
& \textbf{0.69} \textcolor[HTML]{0f98b0}{\scriptsize (+0.07GB)} & \bf{372.25} & \textbf{5.19} & \textbf{746.03} & \textbf{1.01} \textcolor[HTML]{0f98b0}{\scriptsize (+2.0\%)}\\
\bottomrule
\end{tabular}
}
\label{tab:cache-macs-latency}
{\scriptsize
\begin{itemize}[leftmargin=10pt,topsep=0pt]
    \item \textbf{Note:} All methods inherit baseline memory optimizations (e.g., FlashAttention). ToCa is incompatible with these, hence its higher reported cache usage.  \textbf{Actual cache memory usage} for each method should be computed as: \texttt{VRAM Overhead $-$ 0.62\,GB}.
\end{itemize}
}
\vspace{-2mm}

\end{table}

\subsubsection{Decomposition and Order of Prediction Ablation Study}
\begin{figure}[htbp]
    \centering
    \vspace{-3mm}
    \begin{minipage}[t]{0.57\linewidth}
         \centering
        \raisebox{4mm}{\includegraphics[width=\linewidth]{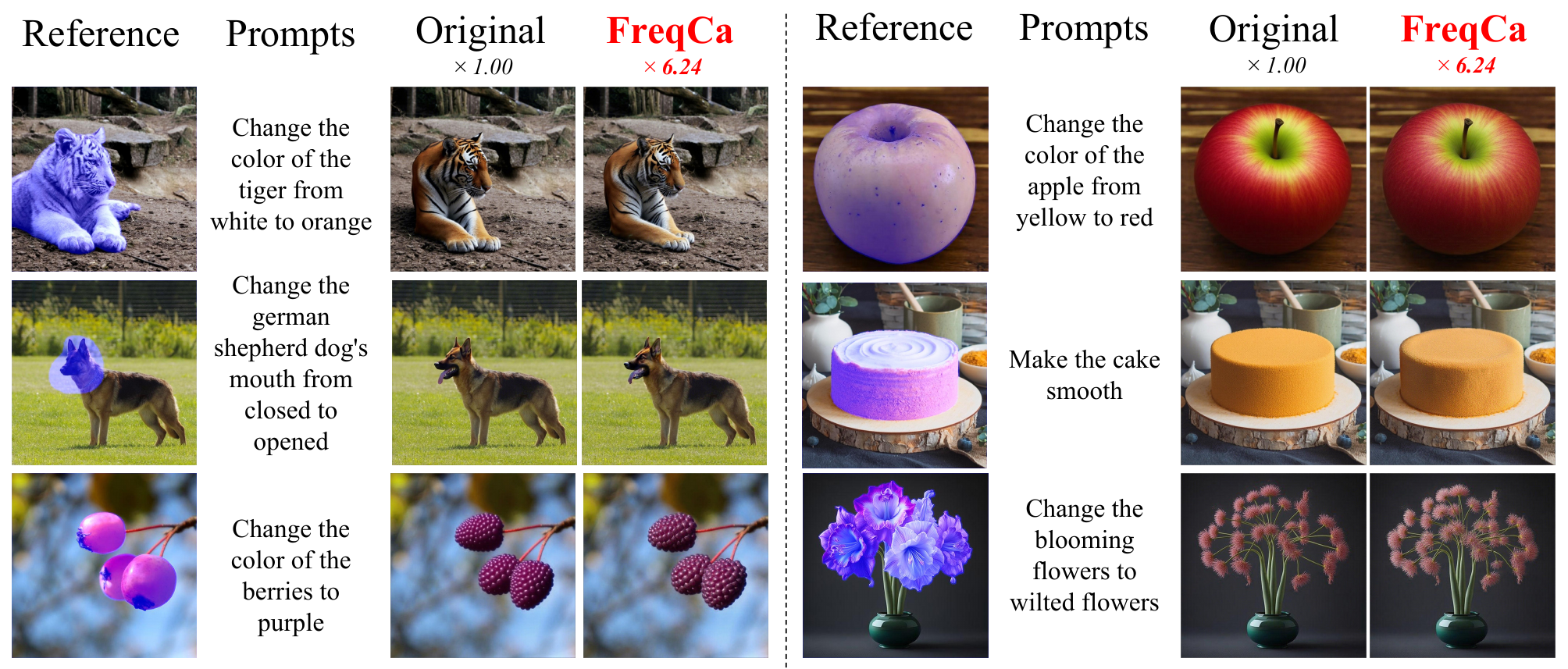}}
        \vspace{-6mm}
        \caption{On FLUX.1-Fill-dev, \textit{FreqCa} achieves 6.24$\times$ acceleration while preserving image inpainting quality indistinguishable from the original.}
        \label{fig:flux-fill}
        \vspace{-5mm}
    \end{minipage}
    \hfill
    \vspace{-3mm}
    \begin{minipage}[t]{0.4\linewidth}
        \centering
        \includegraphics[width=1\linewidth]{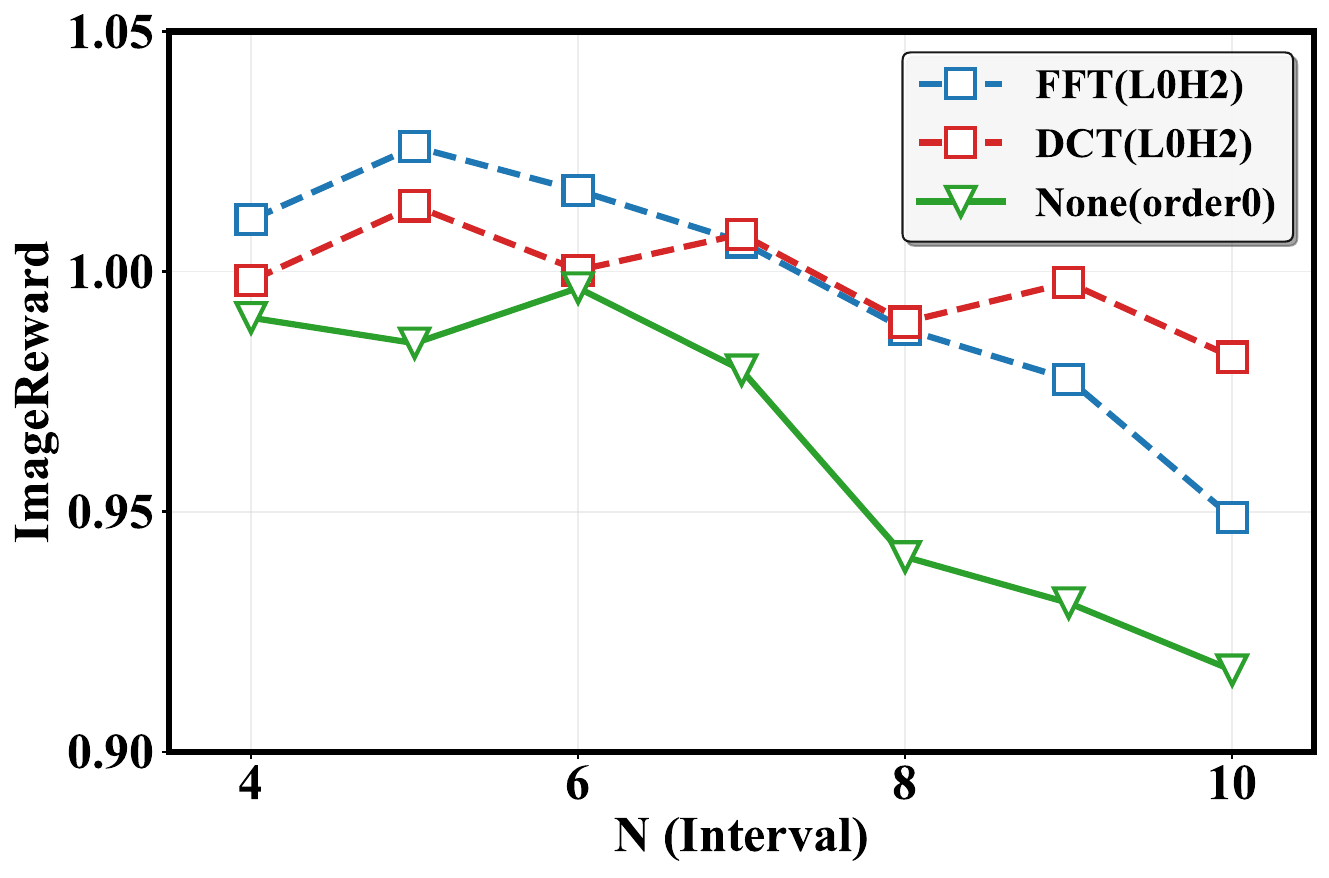}
        \vspace{-6mm}
        \caption{Comparing optimal predictors under varied frequency decompositions on FLUX across speedup ratios.}
        \label{fig:flux-ablation-single}
        \vspace{-5mm}
    \end{minipage}
\end{figure}

\vspace{-1.5mm}
We perform an ablation study on FLUX.1-dev to identify the optimal frequency decomposition method and prediction order. The study compares three decomposition strategies including FFT, DCT, and a baseline without decomposition, each paired with various prediction approaches for frequency components. Figure~\ref{fig:flux-ablation-single} compares these optimized configurations. The results indicate that the DCT-based approach, particularly with low-frequency reuse and high-frequency prediction, achieves consistently high ImageReward across acceleration ratios and demonstrates marked superiority at larger intervals ($N > 8$). This robustness under high acceleration factors confirms the rationale for our choice. A separate ablation on Qwen-Image revealed that the FFT-based method with the same prediction strategy performed best. As shown in Figure~\ref{fig:qwen-image-ablation}, other configurations result in significant quality degradation when compared to our optimal settings.

\section {Conclusion}
\vspace{-3mm}

In this work, we presented \textbf{\textit{FreqCa}}, a frequency-aware feature caching framework that unifies the strengths of reuse- and forecast-based paradigms. By decomposing features into low- and high-frequency components, \textit{FreqCa} selectively reuses stable low-frequency features and accurately predicts dynamic high-frequency components, leading to a superior trade-off between acceleration and generation quality. Furthermore, by introducing Cumulative Residual Feature caching, we reduced the memory footprint to $\mathcal{O}(1)$, making frequency-aware caching practical even on consumer hardware. Extensive experiments across diverse diffusion models demonstrate that \textit{FreqCa} achieves 6–7$\times$ acceleration with negligible quality degradation, establishing a new SOTA in efficient diffusion inference. We believe \textit{FreqCa} opens up new possibilities for scalable, high-performance generative modeling and offers a general method for future research in frequency-aware acceleration techniques.

\section*{Ethics Statement}
\vspace{-3mm}
This work presents FreqCa, a technical method for accelerating diffusion model inference through frequency-aware caching. Our research focuses purely on computational efficiency improvements and does not introduce new ethical risks beyond those inherent to the underlying diffusion models. We use only publicly available models and datasets in our experiments, and our acceleration technique is model-agnostic and content-neutral. While our method reduces inference time and computational costs, which could potentially make generative AI more accessible, we acknowledge that this accessibility applies to both beneficial and potentially harmful use cases. We encourage responsible deployment of accelerated diffusion models in accordance with existing ethical guidelines for AI-generated content, including proper disclosure of synthetic media and consideration of potential societal impacts.

\section*{Reproducibility Statement}
\vspace{-3mm}
We are committed to ensuring the full reproducibility of our FreqCa framework. To this end, Section 3 provides the complete mathematical formulations for our core algorithmic components: frequency decomposition (FFT/DCT), Cumulative Residual Feature (CRF) caching, and second-order Hermite prediction. All experimental configurations are detailed in Section 4.1, specifying the models evaluated (e.g., FLUX.1-dev, Qwen-Image), the datasets used (DrawBench and GEdit), and the full set of evaluation metrics (e.g., ImageReward, CLIP Score, PSNR). Section 4 and the Appendix present our detailed ablation studies, hyperparameter choices for decomposition methods and prediction strategies, and the computational complexity analysis. An anonymous source code repository is provided in the supplementary materials, containing complete inference and training scripts, configuration files with random seeds, and data preprocessing pipelines. The repository will be made publicly available upon acceptance.

\bibliography{iclr2026_conference}
\bibliographystyle{iclr2026_conference}

\appendix
\newpage

\renewcommand{\thefigure}{\Alph{section}\arabic{figure}} 
\renewcommand{\thetable}{\Alph{section}\arabic{table}}   
\setcounter{figure}{0}
\setcounter{table}{0}
\section*{APPENDIX}

\section{Use of Large Language Models}

No Large Language Models were used in this research. All research ideas, algorithmic designs, experimental methodologies, data analysis, and manuscript writing were entirely completed by the authors independently.



\section{Detailed Experimental Setup}
\label{appendix:experiment_setup}

This section provides comprehensive technical details for all experimental configurations mentioned in Section 4.1.

\subsection{Model and Task Specifications}
\paragraph{$\text{FLUX}$.1-dev and Qwen-Image:}
The images generated by the FLUX.1-dev, FLUX.1-schnell, and Qwen-Image-Lightning models are obtained at 1024x1024 resolution using 200 high-quality prompts sourced from the DrawBench benchmark, while those generated by Qwen-Image were obtained at 1328 $\times$ 1328 resolutions. Quality assessment is performed using ImageReward, a robust perceptual metric for text-to-image alignment.

\paragraph{$\text{FLUX}$.1-Kontext-dev and Qwen-Image-Edit:}
We employ the $\text{FLUX}$.1-Kontext-dev and Qwen-Image-Edit model for image editing synthesis. Images editing and quality assessment is performed using GEdit benchmark, which grounded in real-world usages is developed to support more authentic and comprehensive evaluation of image editing models.

\subsection{Hardware and Computational Resources}

All experiments are conducted on enterprise-grade GPU infrastructure:
\begin{itemize}[leftmargin=10pt,topsep=-2pt]
    \item $\text{FLUX}$.1-dev experiments: NVIDIA A100 GPU
    \item $\text{FLUX}$.1-Kontext-dev experiments: NVIDIA A100 GPU
    \item $\text{Qwen}$-Image experiments: NVIDIA H20 GPU  
    \item $\text{Qwen}$-Image-Edit experiments: NVIDIA H20 GPU  

\end{itemize}

\subsection{FreqCa Implementation Parameters}
\begin{itemize}[leftmargin=10pt,topsep=-2pt]
    \item $\text{FLUX}$.1-dev experiments: DCT-based frequency decomposition was adopted.
    \item $\text{FLUX}$.1-Kontext-dev experiments: DCT-based frequency decomposition was adopted.
    \item $\text{Qwen}$-Image experiments: FFT-based frequency decomposition was adopted.
    \item $\text{Qwen}$-Image-Edit experiments: FFT-based frequency decomposition was adopted.
\end{itemize}

\section{Decomposition and Order of Prediction Ablation Study}

\begin{figure}[htp]
    \centering
    \includegraphics[width=1\linewidth]{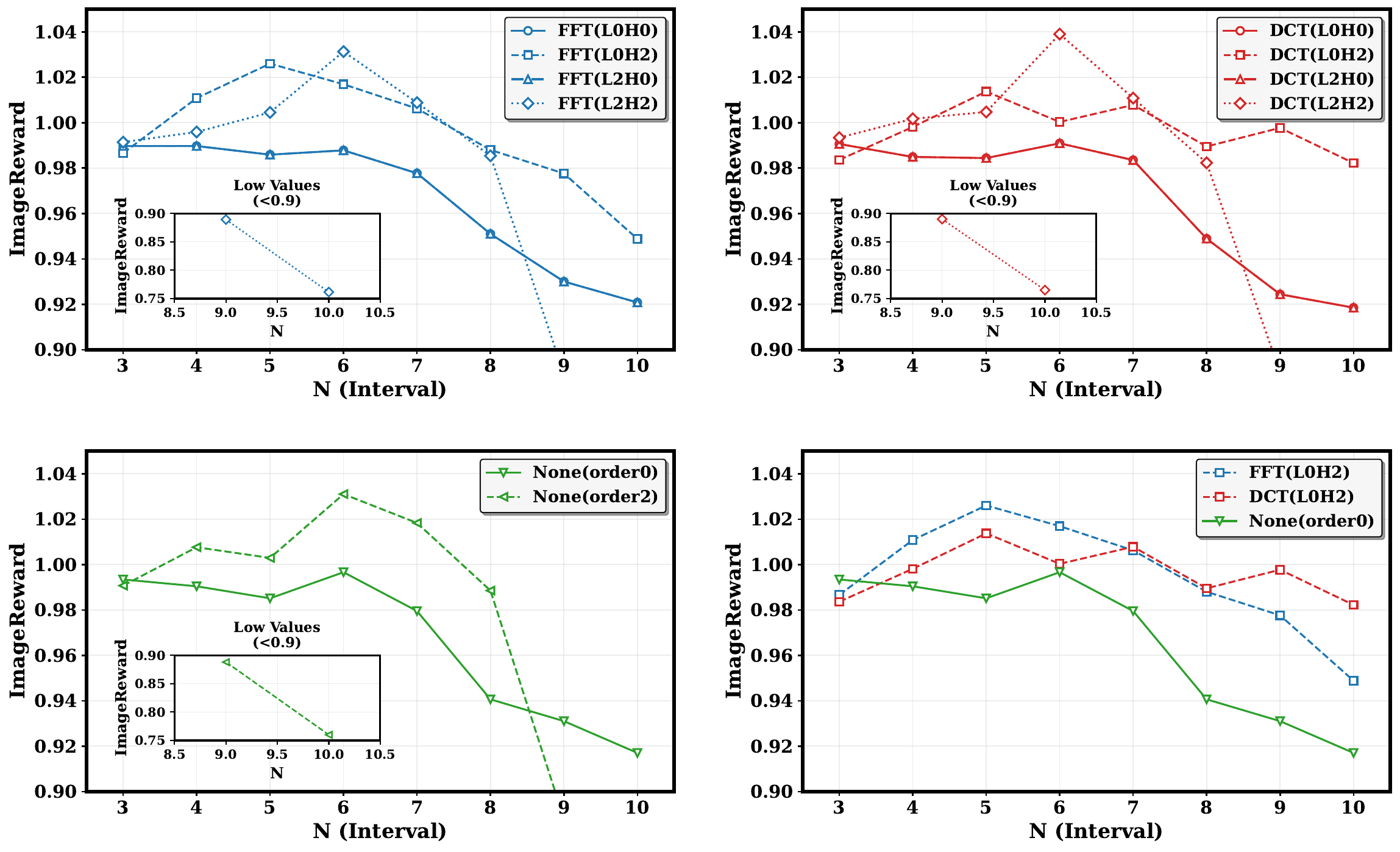}
    \caption{The ImageReward scores of FLUX.1-dev’s decomposition strategies (FFT, DCT, no-decomposition) paired with different frequency prediction approaches are presented here. This content includes the optimal prediction method for each decomposition strategy where low-frequency reuse and high-frequency prediction apply to FFT and DCT, and direct reuse applies to the no-decomposition (None) strategy.}
    \label{fig:flux-order-combination}
\end{figure}

As shown in Figure~\ref{fig:flux-order-combination}, we systematically compared classical frequency decomposition methods (FFT and DCT) with a baseline that does not perform any frequency decomposition. The results clearly demonstrate that frequency decomposition plays a critical role in stabilizing model performance: omitting decomposition leads to a sharp drop in ImageReward scores, whereas both FFT and DCT significantly mitigate this degradation and maintain stable quality across timesteps. Furthermore, we investigated the impact of different prediction orders for low- and high-frequency components. We observe that inappropriate prediction strategies can easily introduce errors and harm generation quality. Among all tested configurations, the combination of zeroth-order prediction (direct reuse) for low-frequency components and second-order prediction for high-frequency components achieves consistently superior results, validating our hypothesis that low-frequency features should be reused directly while high-frequency components benefit from higher-order modeling. These findings confirm the necessity of frequency-aware design and provide empirical guidance for selecting optimal prediction strategies.

\subsection{Prediction Order Combinations}

\end{document}